\documentclass[preprint,12pt,authoryear]{elsarticle}

\usepackage{amssymb}
\usepackage{amsmath}
\usepackage{hyperref}
\usepackage{xcolor}
\hypersetup{
    colorlinks=true,   
    linkcolor=blue,    
    citecolor=blue,    
    urlcolor=blue      
}
\usepackage{booktabs}
\usepackage{rotating}
\usepackage{amsfonts}
\usepackage[utf8]{inputenc} 
\usepackage[T1]{fontenc}    
\usepackage{multirow}
\usepackage{float}
\usepackage{rotating}
\usepackage{svg}
\usepackage{standalone}
\usepackage{tikz}
\usetikzlibrary{positioning,arrows.meta,shapes.geometric,fit,backgrounds,calc}

\begin{document}

\begin{frontmatter}

\title{Calibrated Uncertainty for Wall-to-Wall Biomass Mapping with Neural Processes}

\author[cam]{Robin Young}
\ead{ray25@cam.ac.uk}

\author[cam]{Srinivasan Keshav}
\ead{sk818@cam.ac.uk}

\affiliation[cam]{organization={Department of Computer Science and Technology},
            institution={University of Cambridge},
            city={Cambridge},
            postcode={CB3 0FD}, 
            state={Cambridgeshire},
            country={UK}}

\begin{abstract}
Wall-to-wall biomass density estimation from NASA's GEDI mission requires interpolating sparse LIDAR observations across heterogeneous landscapes. While Random Forest and XGBoost are widely used, they predict each location independently from satellite features, without adapting to the varying difficulty of heterogeneous landscapes. We introduce Attentive Neural Processes (ANPs), a probabilistic meta-learning architecture that explicitly conditions predictions on local observation sets as well as geospatial foundation model embeddings. ANPs learn a flexible spatial covariance approximator that is more uncertain in complex landscapes and less in homogeneous areas. We validate this approach across five distinct biomes ranging from tropical Amazonian forests to boreal, temperate, and alpine ecosystems, demonstrating that ANPs achieve competitive accuracy while maintaining uncertainty calibration. We demonstrate a scalable, empirically calibrated alternative to ensemble architectures for continental-scale biomass mapping.
\end{abstract}




\end{frontmatter}



\section{Introduction}
\label{intro}

Forest aboveground biomass density (AGBD) is important for carbon accounting, climate change mitigation, and conservation planning \citep{Dubayah2022GEDILA, Duncanson2022GEDI_AGB, Duncanson2023ProtectedAreas, Santoro2024DesignAP}. Accurate and high resolution biomass maps enable monitoring of deforestation, forest degradation, and carbon sequestration which are essential for implementing conservation programs and climate targets. NASA's Global Ecosystem Dynamics Investigation (GEDI) mission provides high quality LIDAR measurements of forest structure, including calibrated estimates of AGBD at $\sim$25m footprint resolution. However, GEDI samples sparsely, with footprints separated by approximately 600m along-track and 60m across-track orbital patterns \citep{Dubayah2020GEDI}. To support operational forest monitoring, these sparse observations must be interpolated to create wall-to-wall maps.

Geospatial foundation models, which are large scale pretrained models that learn rich representations from satellite imagery \citep{feng2025tessera, zhu2024foundationsearthclimatefoundation, huang2025surveyremotesensingfoundation}, provides an opportunity to improve feature quality for mapping tasks. Existing approaches to wall-to-wall biomass mapping from GEDI data typically aggregate shots to coarse resolution ($\sim$1km) to increase sampling density \citep{Duncanson2022GEDI_AGB, Dubayah2022GEDILA}. Recently, machine learning and deep learning methods have enabled finer-resolution mapping (10-30m) by combining GEDI footprints with multispectral optical and synthetic aperture radar (SAR) imagery \citep{Shendryk2022_FusingGEDI, sialelli2025agbd, nascetti2023biomassters, Pascarella2023ReUse}. These approaches typically use ensemble methods such as Random Forest \citep{breiman2001random} or XGBoost \citep{Chen2016xgboost} to predict biomass from satellite derived features at arbitrary locations. Separately, encoder-decoder architectures including U-Nets and Vision Transformers have been applied to dense biomass prediction from gridded satellite imagery \citep{Lang2022GlobalCanopyHeight, nascetti2023biomassters}, where both inputs and outputs are spatially continuous rasters.

Biomass point estimates alone are insufficient for decision making because uncertainty quantification is essential \citep{Araza2022ACF}. Conservation planners need to know where estimates are reliable, to prioritize field validation efforts and assess risk \citep{Song2023}. Carbon accounting also requires quantifying uncertainty to meet reporting standards \citep{iso14064-1-2018}. In particular, carbon accounting frameworks such as the Verified Carbon Standard and ART TREES require that map-derived biomass estimates be validated and that their uncertainty be quantified and propagated into reported confidence intervals. Reliable pixel-level uncertainty is a prerequisite for these validation and propagation steps. Despite this need, most biomass mapping studies provide only limited uncertainty representations, typically reporting cross-validation RMSE or ensemble variance as proxies for predictive uncertainty \citep{Li2020, Xu2024ForestAB, Zurqani2025AMA, Turton2022Improving, Li2022RF_MC, Yang2023China, Liu2024Shrub}. Evaluations focused on prediction interval calibration remain rare in GEDI-based biomass mapping. When biomass mapping studies report uncertainty, do these uncertainties actually reflect the data?

This gap is noteworthy for GEDI-based biomass mapping due to the mission's irregular sampling pattern, and more broadly for ecological modeling where sparse but clustered sampling is the norm. Recent work has validated prediction interval coverage for GLM-based biomass estimation using airborne LIDAR \citep{mukhopadhyay2024computation}, but equivalent scrutiny has not been applied to the ML ensemble methods standard in GEDI wall-to-wall mapping. Intuitively, uncertainty should depend on context. Predictions for locations far from GEDI footprints should be more uncertain than those nearby, and predictions in heterogeneous landscapes should be more uncertain than in homogeneous areas. Ensemble methods can incorporate spatial coordinates as features but do not condition on the set of nearby observations, limiting their ability to adapt uncertainty to local spatial context. Recent work has explored Bayesian and hierarchical approaches to GEDI-based biomass interpolation \citep{May2024_Indonesia_AGBD}, and applied deep ensembles to estimate canopy heights \citep{Lang2022GlobalCanopyHeight}.

We introduce an alternative approach based on Neural Processes (NPs) \citep{garnelo2018conditional, kim2018attentive}, a class of models that explicitly condition predictions on nearby observed data points and naturally produce calibrated uncertainty estimates. Gaussian Processes \citep{Rasmussen2005GP} provides a theoretically sound framework for this problem, but it scales cubically with the number of data points, making it operationally difficult for GEDI-scale datasets. Neural processes offer a way to capture the spirit of GP-style spatial conditioning in a scalable, amortized fashion. Neural processes treat any set of context observations $(x_c, y_c)$ as conditioning information, learn a distribution over functions given those contexts, and make predictions at arbitrary target locations $x_t$, which is aligned with the GEDI mapping problem. We combine NPs with embeddings from a remote sensing foundation model trained on Sentinel-1 and Sentinel-2 data \citep{feng2025tessera}, demonstrating that this architecture can effectively map GEDI L4A biomass estimates to wall-to-wall predictions with well-calibrated uncertainty.

We evaluate uncertainty calibration for GEDI-based biomass mapping, revealing that standard tree or gradient boosting methods exhibit poor calibration. Then, we show that Attentive Neural Processes are competitive in accuracy while maintaining well calibrated uncertainty. We validate these findings across five diverse regions spanning temperate (Maine, US), alpine (South Tyrol, Italy), boreal (Hokkaido, Japan), tropical lowlands (Guaviare, Colombia), and tropical montane (Tolima, Colombia) forests. The poor calibration of standard methods suggests that reported uncertainties may be unreliable, potentially leading to misguided conservation decisions or inaccurate carbon accounting. We provide a empirically calibrated and scalable alternative for biomass mapping with quantified uncertainty.

\section{Methods}
\label{methods}

\subsection{Data Sources and Preprocessing}

\subsubsection{GEDI L4A Aboveground Biomass Density}

We obtain GEDI Level 4A (L4A) V2 aboveground biomass density (AGBD) estimates from a single year using the gediDB library \citep{Besnard2025}, which provides efficient spatial queries. To facilitate alignment with satellite imagery and avoid spatial autocorrelation during validation, we organize L4A shots into a tile structure with 0.1$^{\circ}$ $\times$ 0.1$^{\circ}$ geographic tiles.

We apply quality filtering based on beam sensitivity, surface detection quality, and agreement with elevation data (detailed criteria in Table~\ref{tab:gedi_filtering} and \ref{app:preprocessing}) to exclude low confidence estimates. Following standard practice for biomass modeling \citep{Chave2014}, we apply log transformation to stabilize variance and address the log-normal distribution of biomass data. Additionally, we filter out AGBD values over 500 Mg/ha following prior works \citep{sialelli2025agbd, Carreiras2017Coverage}. Values exceeding this threshold (representing <1\% of observations) are rare and likely reflect instrumental artifacts rather than true biomass, which are occasionally present in the GEDI L4A product even after quality filtering. This threshold could in principle remove valid extreme biomass values alongside artifacts, potentially underestimating uncertainty in the highest biomass stands. However, the majority of filtered observations originate from mountainous regions (South Tyrol: 0.79\%, Tolima: 0.96\%), consistent with known GEDI waveform processing artifacts on steep slopes (Table~\ref{tab:filtering_sensitivity}). GEDI L4A AGBD values are themselves estimates with associated uncertainty \citep{Dubayah2022GEDILA}. In this work, we treat GEDI estimates as the target variable.

\subsubsection{Remote Sensing Foundation Model Embeddings}

Rather than using hand crafted spectral indices or raw satellite imagery, we use embeddings from Tessera \citep{feng2025tessera}, a transformer-based remote sensing foundation model pretrained on Sentinel-1 (C-band SAR) and Sentinel-2 (multispectral optical) imagery at global scale. We retrieve embeddings matched to the GEDI shots via the geotessera Python package \citep{feng2025tessera} and use them as frozen features. Tessera ingests time series of ten Sentinel-2 spectral bands (B2--B8, B8A, B11, B12) and both Sentinel-1 polarizations (VV, VH), producing 128-dimensional embeddings at 10m resolution that integrate spectral, temporal, and structural information.

For each GEDI footprint location, we extract a 3$\times$3 pixel patch (30m$\times$30m spatial context) of embeddings centered on the footprint coordinates. This patch size captures local neighborhood characteristics while remaining computationally efficient. The context window is comparable to the GEDI footprint diameter (25m) and provides information about the immediate surrounding landscape that may influence biomass estimates.

We normalize spatial coordinates to the [0,1] range using global bounds from the study area:
\begin{equation}
    \text{coord}_{\text{norm}} = \frac{\text{coord} - \text{coord}_{\min}}{\text{coord}_{\max} - \text{coord}_{\min}}
\end{equation}
During training only, we add Gaussian noise to the normalized coordinates. This makes the model robust to slight GPS errors and prevents it from overfitting to exact coordinate locations \citep{Roy2021GEDI}.

For AGBD values, we apply a normalized log transformation:
\begin{equation}
    \text{AGBD}_{\text{norm}} = \frac{\log(1 + \text{AGBD})}{\log(1 + 200)}
\end{equation}
which compresses the range while maintaining interpretability, with the denominator chosen to map the effective dynamic range of biomass (0-200 Mg/ha) to the [0,1] interval, ensuring that the model maintains gradient sensitivity in the regime where the inputs are most informative. This transformation is inverted during inference to report predictions in original Mg/ha units.

\subsection{Neural Process Architecture}

We implement an Attentive Neural Process (ANP) \citep{kim2018attentive}, which extend the Conditional Neural Process model \citep{garnelo2018conditional} with attention mechanisms for improved context aggregation. Neural processes are meta-learning models that learn distributions over functions by conditioning on a set of observed context points. Unlike standard supervised learning methods that learn a single function mapping inputs to outputs, NPs learn to produce a different predictive distribution for each set of context observations.

\subsubsection{Architecture Components}

The ANP architecture consists of five main components. A detailed specification of each component's architecture is provided in \ref{app:architecture}.

\textbf{Embedding encoder}: A 3-layer convolutional neural network (CNN) with residual connections processes each 3$\times$3$\times$128 embedding patch into a 1024-dimensional feature vector. This encoder captures spatial patterns within the local neighborhood around each GEDI footprint.

\textbf{Context encoder}: A 3-layer multilayer perceptron (MLP) with layer normalization encodes each context observation consisting of the feature vector, normalized coordinates, and observed AGBD value into a representation.

\textbf{Deterministic path}: Multihead cross attention with 16 heads aggregates information from context points to each target location. The attention mechanism weights contributions from context points based on their relevance to the target, encoding spatial relationships such as distance and similarity.

\textbf{Stochastic latent path}: Mean pooled context representations parameterize a latent distribution $\mathcal{N}(\mu_z, \sigma_z)$, which captures global uncertainty about the underlying function. Samples from this distribution (via the reparameterization trick \citep{kingma2014vae}) are combined with the deterministic path to form the final representation.

\textbf{Decoder}: An MLP outputs parameters of a Gaussian predictive distribution $\mathcal{N}(\mu, \sigma^2)$ for each target location. The mean $\mu$ provides the point estimate, while the variance $\sigma^2$ quantifies predictive uncertainty.

The separation into deterministic and stochastic paths allows the model to capture both function specific structure (via attention to context) and global uncertainty (via the latent variable), providing better uncertainty estimates than single path architectures  (see \ref{appendix:ablation} for ablation). Intuitively, the deterministic path acts like a learned interpolation, allowing the model to smooth biomass values based on feature similarity. The stochastic path estimates the global uncertainty for the current landscape, increasing variance when the context points (GEDI shots) are sparse or contradictory.

Predicting location-specific $\sigma^2$ via NLL minimization is not unique to NPs, and any neural network can be trained this way. The difference here is that this prediction is conditioned not only on the target location's features but on the set of values from observed context points, enabling uncertainty that adapts to local observation density and spatial heterogeneity rather than reflecting only feature-space properties.

\subsubsection{Training Objective}

Neural processes are trained using amortized inference, which learns an approximate distribution over functions by optimizing a tractable lower bound on the data likelihood. The model is trained to maximize the evidence lower bound (ELBO):
\begin{equation}
    \mathcal{L} = -\mathbb{E}_{q(z|C,T)}[\log p(y_t | x_t, z, C)] + \beta \cdot \text{KL}[q(z|C, T) \| p(z|C)]
\end{equation}
where $C = \{(x_c, y_c)\}$ denotes context observations, $T = \{(x_t, y_t)\}$ denotes target observations, and $q(z|C,T)$ is the approximate posterior over latent variables. The first term is the negative log-likelihood (NLL) of target observations:
\begin{equation}
    \text{NLL} = \frac{1}{2}\left[\log(\sigma^2) + \frac{(y - \mu)^2}{\sigma^2}\right]
\end{equation}
which encourages accurate predictions with appropriately scaled uncertainties. The second term is the Kullback-Leibler (KL) divergence between the approximate posterior and the learned prior, which regularizes the latent space. We use $\beta$-VAE warmup \citep{higgins2017betavae} to prevent posterior collapse, linearly increasing $\beta$ over the first 10 epochs. This allows the model to first learn useful representations before regularization.

\subsection{Training Procedure}

We train ANPs using the AdamW optimizer \citep{loshchilov2018decoupled} with learning rate $5 \times 10^{-4}$, gradient clipping (max norm 1.0), and a scheduler that reduces learning rate by factor 0.5 after 5 epochs without validation improvement. Training continues until validation loss fails to improve for 15 consecutive epochs (early stopping).

Following the neural process modeling framework \citep{garnelo2018conditional, kim2018attentive}, training is done episodically. During each training iteration, we sample a tile, randomly partition observations within each tile into context and target sets, with the context ratio sampled uniformly from [0.3, 0.7], and computes the loss by comparing predicted target distributions to GEDI observations. This variable sized context encourages the model to learn robust function representations that generalize across different numbers of observations, which is a form of meta-learning.

For robust evaluation, we train each model configuration across 10 random seeds and report mean $\pm$ standard deviation for all metrics. To generate wall-to-wall maps, we iterate through each tile in the region of interest. Complete training hyperparameters are provided in \ref{app:implementation}.

\subsection{Spatial Cross-Validation}

Standard random train-test splits are inappropriate for spatial data due to spatial autocorrelation, which can inflate performance estimates if nearby observations appear in training and test sets \citep{Roberts2017, ploton2020spatial}. We use buffered spatial cross-validation for evaluation.

We partition geographic tiles (0.1$^{\circ}$ $\times$ 0.1$^{\circ}$) into training (70\%), validation (15\%), and test (15\%) sets. We also exclude any training tiles next to test tiles. This buffer is sufficient to mitigate the majority of spatial autocorrelation effects in forest biomass for typical environments \citep{RejouMechain2014, OtavianoDeAlmeida2025}.

\subsection{Baseline Models}

We compare ANP against four baseline approaches representing common practices in biomass mapping:

\textbf{Random Forest} (RF): We use Random Forest with ensemble sizes ranging from 50-1000 trees and maximum depths from 1-20. Uncertainty estimates are derived from the variance of predictions across trees as a simple baseline method. While computationally efficient and widely used, RF uncertainty estimates only reflect ensemble variance (disagreement between trees) rather than true predictive uncertainty \citep{Mentch2016rf}.

\textbf{XGBoost}: We use gradient boosting with quantile regression loss to estimate percentiles of the conditional distribution, approximating $\pm 1\sigma$ intervals assuming normality, alongside the mean estimate, as a more principled baseline of uncertainty quantification compared to ensemble variance \citep{koenker1978regression}. Hyperparameters range from tree depths 1-20 and boosting from 50-1000. Quantile regression provides explicit uncertainty estimates but fits global residual distributions.

\textbf{MLP with MC Dropout}: We train a 3-layer MLP (512-256-128 units) with dropout rate 0.2 and estimate uncertainty via 100 Monte Carlo forward passes at test time \citep{Gal2016mcdropout}. MC dropout approximates Bayesian inference and provides uncertainty estimates but is computationally expensive at test time from multiple forward passes and may underestimate uncertainty in regions far from training data.

\textbf{Inverse Distance Weighting} (IDW): We use this as a spatial interpolation baseline that ignores satellite features, providing a lower bound on performance. IDW predictions are computed as weighted averages of nearby GEDI observations with weights $w_i = 1/d_i^2$, where $d_i$ is the distance to observation $i$. This baseline tests whether learned models trained on embeddings provide value beyond simple spatial interpolation.

All learning based baselines use the same input features of concatenated normalized coordinates and flattened embedding patches. We perform hyperparameter sweeps for RF and XGBoost (60 configurations each) for a fair comparison with the ANP. We additionally discuss results for other baseline models in \ref{appendix:kriging}, \ref{appendix:qrf}, and \ref{appendix:deep_ensemble}.

\subsection{Evaluation Metrics}

We evaluate models along two dimensions, predictive accuracy and uncertainty calibration.

\subsubsection{Accuracy Metrics}

We compute performance metrics in log-transformed space to match the training objective, then report additional metrics after back transformation to original Mg/ha units for interpretability:
\begin{itemize}
    \item Log-space $R^2$: Coefficient of determination in log-space, our primary accuracy metric
    \item Log-space RMSE and MAE: Root mean squared error and mean absolute error in log-space
    \item Linear-space RMSE and MAE: Errors in original Mg/ha units after back transformation for reference
\end{itemize}

We do not report linear-space $R^2$ because the exponential back-transformation induces heteroscedastic residuals, violating the assumptions under which $R^2$ is interpretable \citep{Baskerville1972}.

\subsubsection{Calibration and Bias Metrics}

Total uncertainty in map-based biomass estimation comprises several components: measurement uncertainty in the reference observations, model uncertainty in the learned feature-biomass relationship, and spatial/residual uncertainty arising from prediction away from observations \citep{Araza2022ACF}. This work focuses on producing calibrated predictive distributions at the pixel level, that is, capturing model and spatial-residual uncertainty conditional on the GEDI reference, and does not attempt to propagate reference--measurement uncertainty from the GEDI L4A product itself, which we treat as an isolated target variable.

Prediction intervals quantify uncertainty about individual future observations while ensemble variance measures disagreement between models fit on different data samples \citep{Kuleshov2018accurate}. The latter conflates sampling variability with predictive uncertainty and fails to adapt to spatial context. Most biomass mapping studies either propagate measurement uncertainty or report modeling ensemble variance or cross-validation RMSE as uncertainty proxies but do not assess whether these provide calibrated prediction intervals for a given point \citep{Turton2022Improving, Araza2022ACF}. We assess calibration using metrics that test whether predicted uncertainties match empirical error distributions:

\textbf{Z-score statistics}: We compute standardized residuals:
\begin{equation}
    z = \frac{y_{\text{true}} - y_{\text{pred}}}{\sigma_{\text{pred}}}
\end{equation}
If uncertainties are well-calibrated, $z$ should follow $\mathcal{N}(0, 1)$. We report:
\begin{itemize}
    \item \textbf{Z-score mean} (ideal: 0.0) indicates systematic bias in uncertainty estimates
    \item \textbf{Z-score standard deviation} (ideal: 1.0) indicates calibration of uncertainty scale. Values $> 1$ indicate overconfident (too narrow) predictions. Values $< 1$ indicate underconfident (too wide) predictions.
\end{itemize}

\textbf{Coverage statistics}: The fraction of GEDI L4A observations falling within predicted intervals. For Gaussian predictive distributions, we expect 68.27\% within $\pm 1\sigma$, 95.45\% within $\pm 2\sigma$, and 99.73\% within $\pm 3\sigma$. Deviations from these nominal values indicate miscalibration. Coverage below nominal values indicates overconfidence; coverage above nominal values indicates underconfidence.

Together, these metrics distinguish between models that provide accurate point estimates with unreliable uncertainty (poor calibration) versus models that provide trustworthy prediction intervals (good calibration). All calibration and bias metrics are computed in log-transformed space, where the model's Gaussian predictive distribution is defined and the distributional assumptions hold.

\subsection{Study Area}

\subsubsection{Primary Study Region}

Our primary experiments are conducted in the Guaviare department of Colombia (1$^{\circ}$ $\times$ 1$^{\circ}$ region from -73$^{\circ}$W to -72$^{\circ}$W, centered on 2.5$^{\circ}$N), which covers tropical forests experiencing varying degrees of disturbance, deforestation, and regeneration. The region has experienced sustained forest conversion driven by cattle ranching, smallholder agriculture, and historically coca cultivation. Documentation of fine-scale disturbance dynamics is limited by the region's remote and partially contested nature. This region provides diverse biomass conditions ranging from cleared land to intact forest, which allows model evaluation across the full range of interest for conservation and carbon monitoring applications.

\subsubsection{Cross-Regional Generalization}

To assess model robustness across biomes and test spatial extrapolation capabilities, we evaluate performance on four additional regions:

\begin{itemize}
    \item \textbf{Maine, USA} (45.0$^{\circ}$N, 68.5$^{\circ}$W): Temperate mixed forests dominated by conifers and hardwoods, representing northern temperate ecosystems. The region experienced extensive clearing through the 19th century followed by reforestation and is currently characterized by managed second-growth and industrial timberlands, with a comparatively well-documented forestry history through the US Forest Service's Forest Inventory and Analysis program.
    
    \item \textbf{South Tyrol, Italy} (46.5$^{\circ}$N, 11.5$^{\circ}$E): Alpine and subalpine forests at elevations 500-2500m, testing performance in mountainous terrain with steep topographic gradients. The forests have a long established Alpine forestry tradition with relatively stable forest cover in recent decades and active management for both timber and protective functions.
    
    \item \textbf{Hokkaido, Japan} (43.5$^{\circ}$N, 143.0$^{\circ}$E): Boreal and cool-temperate forests, representing high latitude ecosystems with distinct seasonal dynamics. The region's forests include both natural mixed stands and post-Meiji plantation forestry, with native deciduous broadleaves (oak, beech) intermixed with planted conifers (Sakhalin fir, Yezo spruce, larch).
    
    \item \textbf{Tolima, Colombia} (4.5$^{\circ}$N, 75.5$^{\circ}$W): Tropical montane forests in the Andes, as a second tropical site with different topographic and climatic characteristics from Guaviare. Land use is heterogeneous and stratified by elevation, with smallholder coffee cultivation and pasture predominant at lower elevations and intact forest patches concentrated on steep terrain at higher elevations less suitable for agriculture.
\end{itemize}

These regions span tropical, temperate, boreal, and alpine biomes with varying forest structure, composition, species diversity, and biomass distributions. Additional details on all regions, including specific bounding boxes used, are described in \ref{app:region_defs}.

\section{Results}
\label{results}


We begin by evaluating model performance on our primary study region (Guaviare, Colombia) using spatial cross-validation. Table~\ref{tab:summary} presents accuracy and calibration metrics for ANP and representative baseline configurations, averaged over 10 seeds.

\begin{table}[H]
\centering
\caption{Summary of model performance on Guaviare test set. Values are mean $\pm$ standard deviation over 10 seeds. Best model for each metric shown in bold.}
\label{tab:summary}
\hspace*{-1.5cm}
\resizebox{1.25\textwidth}{!}{
\begin{tabular}{lccccc}
\toprule
\textbf{Model} & \textbf{ANP} & \textbf{XGBoost} & \textbf{MLP w/ Dropout} & \textbf{Random Forest} & \textbf{IDW} \\
\midrule
\multicolumn{6}{l}{\textit{Accuracy Metrics}} \\
Validation Log $R^2$       & \textbf{0.735 $\pm$ 0.053} & 0.707 $\pm$ 0.062 & 0.704 $\pm$ 0.061 & 0.692 $\pm$ 0.061 & -0.266 $\pm$ 0.175 \\
Test Log $R^2$             & \textbf{0.747 $\pm$ 0.043} & 0.737 $\pm$ 0.042 & 0.735 $\pm$ 0.046 & 0.724 $\pm$ 0.045 & -0.229 $\pm$ 0.111 \\
Test Log RMSE           & \textbf{0.199 $\pm$ 0.016} & 0.203 $\pm$ 0.014 & 0.203 $\pm$ 0.016 & 0.207 $\pm$ 0.015 & 0.440 $\pm$ 0.022 \\
Test Log MAE            & \textbf{0.141 $\pm$ 0.012} & 0.142 $\pm$ 0.011 & \textbf{0.141 $\pm$ 0.011} & 0.148 $\pm$ 0.010 & 0.361 $\pm$ 0.026 \\
Linear RMSE (Mg/ha)     & 50.56 $\pm$ 5.36 & \textbf{49.35 $\pm$ 2.53} & 50.02 $\pm$ 2.75 & 51.68 $\pm$ 3.52 & 76.27 $\pm$ 6.36 \\
Linear MAE (Mg/ha)      & 27.33 $\pm$ 3.44 & \textbf{26.68 $\pm$ 2.08} & 27.02 $\pm$ 2.27 & 28.01 $\pm$ 2.71 & 50.16 $\pm$ 4.73 \\
\midrule
\multicolumn{6}{l}{\textit{Uncertainty Calibration (Ideal values in parentheses)}} \\
Z-Score Mean (0.0)      & 0.023 $\pm$ 0.124 & \textbf{-0.014 $\pm$ 0.077} & 0.041 $\pm$ 0.249 & -0.109 $\pm$ 0.407 & -- \\
Z-Score Std (1.0)       & \textbf{0.997 $\pm$ 0.099} & 1.408 $\pm$ 0.157 & 4.140 $\pm$ 0.547 & 6.960 $\pm$ 2.065 & -- \\
1$\sigma$ Coverage (>68.3\%) & \textbf{79.1 $\pm$ 2.8} & 65.2 $\pm$ 2.5 & 26.1 $\pm$ 6.0 & 19.1 $\pm$ 3.3 & 44.9 $\pm$ 2.9 \\
2$\sigma$ Coverage (>95.4\%) & \textbf{94.1 $\pm$ 1.4} & 90.0 $\pm$ 1.9 & 53.0 $\pm$ 6.1 & 39.8 $\pm$ 5.0 & 76.9 $\pm$ 4.1 \\
3$\sigma$ Coverage (>99.7\%) & \textbf{98.0 $\pm$ 0.9} & 95.9 $\pm$ 1.3 & 73.1 $\pm$ 3.7 & 59.8 $\pm$ 5.2 & 88.9 $\pm$ 3.1 \\
\bottomrule
\end{tabular}
}
\end{table}

ANP achieves Log $R^2$ of 0.747, competitive with all baseline methods across all hyperparameter configurations tested. The naive spatial interpolation baseline (IDW) achieves negative R$^2$, confirming that learned models provide value over simple distance-weighted averaging. The most significant differences appear in uncertainty calibration metrics. XGBoost shows Z-score standard deviation of 1.408, indicating overconfident uncertainty estimates that are too narrow. Random Forest attains worse calibration, with Z-std of 6.96. XGBoost achieves only 65.2\% coverage at 1$\sigma$ (vs. nominal 68.3\%), while Random Forest achieves just 19.1\%. ANP achieves better performance on both dimensions with Log $R^2$ of 0.747 (higher than any baseline) and Z-std of 0.997 (nearly nominal calibration). The 1$\sigma$ coverage of 79.1\% exceeds the nominal 68.3\%, indicating conservatism. Training time (146s) is comparable with XGBoost's (124s). The near-zero Z-score mean indicates predictions are unbiased relative to GEDI L4A observations.

\subsection{Comparison to Hyperparameter-tuned Baselines}

To investigate the relationship between predictive accuracy, uncertainty calibration, and model configurations, we swept hyperparameters for Random Forest (60 configurations spanning tree depths 1-20 and ensemble sizes 50-1000) and XGBoost (60 configurations with depths 1-20 and boosting rounds 50-1000). Each configuration was evaluated over 5 seeds on the Guaviare test set. Complete results are provided in \ref{appendix:hyperparams}.

Before examining this tradeoff, we note that predictive accuracy and uncertainty calibration are distinct axes of performance, not substitutes. A model may be accurate yet badly calibrated, or well-calibrated yet inaccurate. Our contribution lies on the calibration axis. ANP achieving well-calibrated uncertainty while retaining competitive accuracy is the central result.

\begin{figure}[H]
\centering
\includegraphics[width=\textwidth]{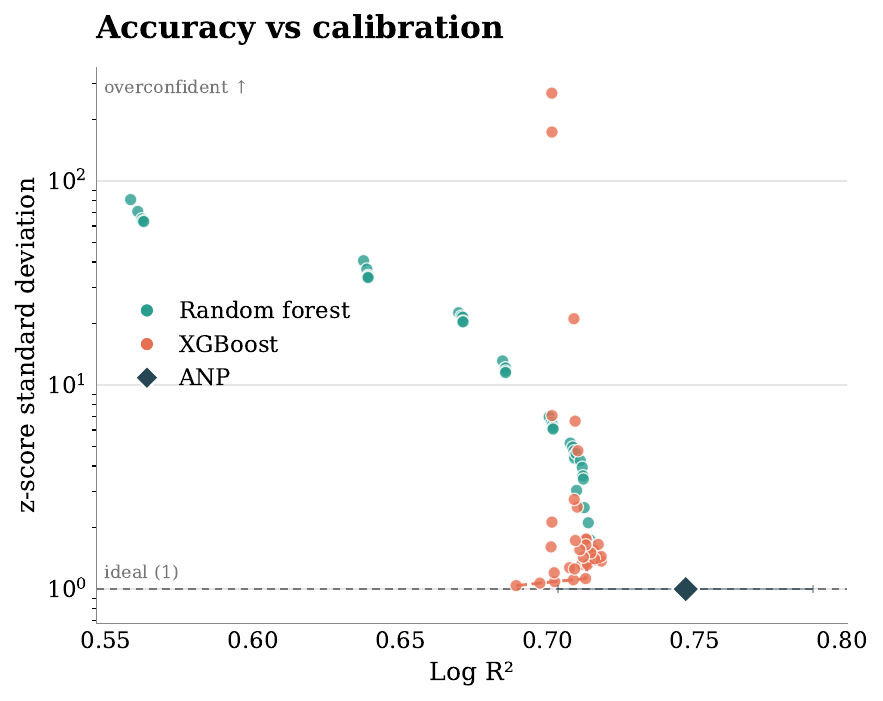}
\caption{Performance in baseline methods. Each point represents a hyperparameter configuration evaluated across 5 random seeds. Horizontal line indicates ideal calibration (Z-score std = 1.0).}
\label{fig:pareto}
\end{figure}

Figure~\ref{fig:pareto} shows that Random Forest shows monotonic improvement in both accuracy and calibration with increasing model complexity. The best RF configuration (depth=20, n=1000) achieves Log $R^2$ of 0.715 with Z-score standard deviation of 1.65, still 65\% overconfident, while requiring 904 seconds training time. Intermediate configurations that researchers might typically use out of the box without explicitly checking calibration (depth=8-10, n=500) show substantial miscalibration (Z-std 4-5, coverage 55-60\%).

The highest-accuracy configurations for XGBoost (depth=6, n=100; Log $R^2$ 0.718) show Z-score standard deviation of 1.44 (44\% overconfident) with only 65.5\% coverage at 1$\sigma$. Configurations approaching ideal calibration (depth=1, n=50; Z-std 1.04) decreases accuracy (Log $R^2$ 0.689). XGBoost performance is scattered with configurations with similar accuracy vary widely in calibration (Z-std 1.1-2.0), and deep models (depth=20) exhibit large miscalibration (Z-std exceeding 100). No configuration simultaneously achieves Log $R^2 > 0.71$ and Z-std $< 1.2$.

ANP (diamond) achieves Log $R^2$ of 0.747 with Z-score standard deviation of 0.997, exceeding both baseline methods on accuracy and calibration simultaneously. Training time of 146 seconds is competitive with XGBoost's best-accuracy configuration (124s) and 6.2$\times$ faster than well-performing RF configurations.

\subsection{Multi-Region Validation}
\label{sec:multiregion}

To assess whether the observed performance patterns generalize beyond our primary study region, we evaluate all models across four ecologically diverse areas spanning tropical, temperate, boreal, and alpine biomes (Table~\ref{tab:regional_results}). Models are trained separately on each region using the same hyperparameters as in the Guaviare analysis, with evaluation repeated over 5 seeds.

\begin{table}[H]
\centering
\caption{Comparison of model performance across four additional testing regions. Values are mean $\pm$ standard deviation.}
\hspace*{-1.5cm}
\label{tab:regional_results}
\resizebox{1.25\textwidth}{!}{
\begin{tabular}{llccccccc}
\toprule
 &  & \multicolumn{2}{c}{\textbf{Accuracy Metrics}} & \multicolumn{2}{c}{\textbf{Z-Score Metrics}} & \multicolumn{3}{c}{\textbf{Coverage Metrics}} \\
\cmidrule(lr){3-4} \cmidrule(lr){5-6} \cmidrule(lr){7-9}
Region & Model & Test Log $R^2$ & RMSE & Mean (0.0) & Std (1.0) & 1$\sigma$ (\%) & 2$\sigma$ (\%) & 3$\sigma$ (\%) \\
\midrule
\multirow{5}{*}{\textbf{Maine, USA}}
& ANP & \textbf{0.581 $\pm$ 0.019} & \textbf{51.37 $\pm$ 2.83} & 0.046 $\pm$ 0.060 & \textbf{0.952 $\pm$ 0.059} & \textbf{75.0 $\pm$ 3.3} & \textbf{95.4 $\pm$ 0.7} & \textbf{99.0 $\pm$ 0.3} \\
& XGBoost & 0.566 $\pm$ 0.020 & 51.59 $\pm$ 2.62 & \textbf{0.020 $\pm$ 0.132} & 1.383 $\pm$ 0.061 & 58.9 $\pm$ 2.0 & 89.1 $\pm$ 0.9 & 97.1 $\pm$ 0.3 \\
& Random Forest & 0.530 $\pm$ 0.023 & 54.53 $\pm$ 2.64 & 0.117 $\pm$ 0.198 & 7.044 $\pm$ 1.810 & 19.4 $\pm$ 2.1 & 38.6 $\pm$ 4.2 & 54.8 $\pm$ 5.6 \\
& MLP w/ Dropout & 0.527 $\pm$ 0.011 & 52.29 $\pm$ 2.20 & 0.384 $\pm$ 0.383 & 4.390 $\pm$ 0.288 & 24.5 $\pm$ 1.3 & 45.8 $\pm$ 1.7 & 61.9 $\pm$ 2.4 \\
& IDW & -0.258 $\pm$ 0.053 & 72.45 $\pm$ 2.27 & -0.058 $\pm$ 0.141 & 2.101 $\pm$ 0.324 & 49.6 $\pm$ 1.7 & 79.6 $\pm$ 2.2 & 91.1 $\pm$ 1.4 \\
\midrule
\multirow{5}{*}{\textbf{South Tyrol, Italy}}
& ANP & \textbf{0.632 $\pm$ 0.012} & 73.05 $\pm$ 2.85 & 0.077 $\pm$ 0.068 & \textbf{1.054 $\pm$ 0.051} & \textbf{69.6 $\pm$ 2.9} & \textbf{94.5 $\pm$ 0.7} & \textbf{98.9 $\pm$ 0.3} \\
& XGBoost & 0.623 $\pm$ 0.013 & \textbf{72.99 $\pm$ 3.78} & \textbf{-0.002 $\pm$ 0.118} & 1.450 $\pm$ 0.163 & 57.8 $\pm$ 1.4 & 88.6 $\pm$ 1.1 & 97.1 $\pm$ 0.9 \\
& Random Forest & 0.566 $\pm$ 0.019 & 79.21 $\pm$ 4.75 & 0.348 $\pm$ 0.405 & 4.649 $\pm$ 0.934 & 24.3 $\pm$ 1.7 & 46.6 $\pm$ 2.5 & 63.5 $\pm$ 3.0 \\
& MLP w/ Dropout & 0.616 $\pm$ 0.010 & 71.95 $\pm$ 3.67 & -0.092 $\pm$ 0.193 & 4.085 $\pm$ 0.312 & 23.8 $\pm$ 1.8 & 45.5 $\pm$ 3.0 & 63.0 $\pm$ 3.2 \\
& IDW & -0.212 $\pm$ 0.176 & 111.11 $\pm$ 8.79 & -0.009 $\pm$ 0.344 & 3.055 $\pm$ 1.920 & 45.4 $\pm$ 2.6 & 75.0 $\pm$ 2.4 & 88.0 $\pm$ 1.9 \\
\midrule
\multirow{5}{*}{\textbf{Hokkaido, Japan}}
& ANP & 0.640 $\pm$ 0.069 & 62.78 $\pm$ 12.16 & \textbf{-0.004 $\pm$ 0.146} & \textbf{0.721 $\pm$ 0.135} & \textbf{86.0 $\pm$ 4.5} & \textbf{97.2 $\pm$ 2.9} & \textbf{99.9 $\pm$ 0.1} \\
& XGBoost & \textbf{0.655 $\pm$ 0.044} & \textbf{60.93 $\pm$ 8.95} & -0.049 $\pm$ 0.197 & 7.686 $\pm$ 11.49 & 56.1 $\pm$ 4.2 & 85.4 $\pm$ 4.2 & 94.3 $\pm$ 2.6 \\
& Random Forest & 0.646 $\pm$ 0.045 & 61.81 $\pm$ 9.78 & 0.032 $\pm$ 0.479 & 6.845 $\pm$ 1.711 & 26.0 $\pm$ 4.0 & 49.0 $\pm$ 6.3 & 64.5 $\pm$ 4.9 \\
& MLP w/ Dropout & 0.626 $\pm$ 0.042 & 62.19 $\pm$ 9.68 & -0.091 $\pm$ 0.412 & 4.365 $\pm$ 0.492 & 26.8 $\pm$ 3.5 & 51.7 $\pm$ 4.8 & 68.7 $\pm$ 5.0 \\
& IDW & -0.121 $\pm$ 0.228 & 86.50 $\pm$ 14.01 & 0.154 $\pm$ 0.712 & 4.775 $\pm$ 2.939 & 44.7 $\pm$ 6.2 & 72.8 $\pm$ 8.9 & 84.7 $\pm$ 6.3 \\
\midrule
\multirow{5}{*}{\textbf{Tolima, Colombia}}
& ANP & 0.583 $\pm$ 0.087 & 79.85 $\pm$ 4.09 & \textbf{-0.013 $\pm$ 0.028} & \textbf{1.081 $\pm$ 0.141} & \textbf{70.8 $\pm$ 5.6} & \textbf{93.0 $\pm$ 3.3} & \textbf{98.1 $\pm$ 1.4} \\
& XGBoost & \textbf{0.598 $\pm$ 0.074} & \textbf{76.25 $\pm$ 4.87} & -0.106 $\pm$ 0.159 & 1.408 $\pm$ 0.116 & 56.7 $\pm$ 2.2 & 87.8 $\pm$ 2.4 & 96.3 $\pm$ 1.4 \\
& Random Forest & 0.551 $\pm$ 0.069 & 80.32 $\pm$ 6.98 & -0.511 $\pm$ 0.440 & 3.856 $\pm$ 0.651 & 24.4 $\pm$ 1.1 & 50.0 $\pm$ 3.8 & 69.3 $\pm$ 4.7 \\
& MLP w/ Dropout & 0.554 $\pm$ 0.070 & 77.13 $\pm$ 5.78 & -0.089 $\pm$ 0.270 & 3.951 $\pm$ 0.343 & 27.5 $\pm$ 2.2 & 53.7 $\pm$ 3.4 & 70.7 $\pm$ 2.7 \\
& IDW & -0.010 $\pm$ 0.159 & 97.41 $\pm$ 7.04 & -0.534 $\pm$ 0.738 & 2.838 $\pm$ 0.875 & 46.1 $\pm$ 1.9 & 78.7 $\pm$ 6.4 & 87.8 $\pm$ 6.3 \\
\bottomrule
\end{tabular}
}
\end{table}

The accuracy-calibration pattern observed in Guaviare holds consistently across all four regions. ANP achieves Z-score standard deviation near 1.0, with 0.95 in Maine, 1.05 in South Tyrol, 0.72 in Hokkaido, and 1.08 in Tolima. Baseline methods show high variance and generally poor calibration. XGBoost Z-std ranges from 1.38 (Maine) to 7.69 (Hokkaido), while Random Forest ranges from 4.65 (South Tyrol) to 7.04 (Maine). MLP with MC Dropout consistently shows poor calibration (Z-std 4.0-4.4) across all regions.

Predictive accuracy shows more variation across regions, likely reflecting differences in forest structural complexity and biomass distributions. In Maine and Tolima, XGBoost achieves slightly higher Log $R^2$ than ANP (0.57 vs 0.58 in Maine, 0.60 vs 0.58 in Tolima), though differences are within one standard deviation. However, XGBoost's accuracy comes with degraded calibration (Z-std 1.38 in Maine, 1.41 in Tolima), yielding coverage of only 57-59\% at 1$\sigma$. ANP maintains 71-75\% coverage in these regions, which is slightly conservative. In South Tyrol, ANP achieves both the highest accuracy (Log $R^2$ 0.63) and best calibration (Z-std 1.05).

Hokkaido shows the highest variance in performance across random seeds (standard deviations 2-3 times larger than other regions), suggesting challenging conditions for biomass estimation. XGBoost achieves marginally higher mean accuracy (Log $R^2$ 0.655 vs 0.640 for ANP), but uncertainty estimates become unstable with Z-std of 7.69 $\pm$ 11.49, indicating that across 5 random seeds, calibration quality varies. This instability manifests in coverage of only 56\% at 1$\sigma$. ANP Z-std decreases to 0.72 (narrower than ideal 1.0, indicating underconfident predictions) and coverage increases to 86\%. Random Forest and MLP show similarly poor calibration (Z-std 6.8 and 4.4 respectively).

\subsection{Calibration Analysis}

\begin{figure}[H]
\centering
\includegraphics[width=\textwidth]{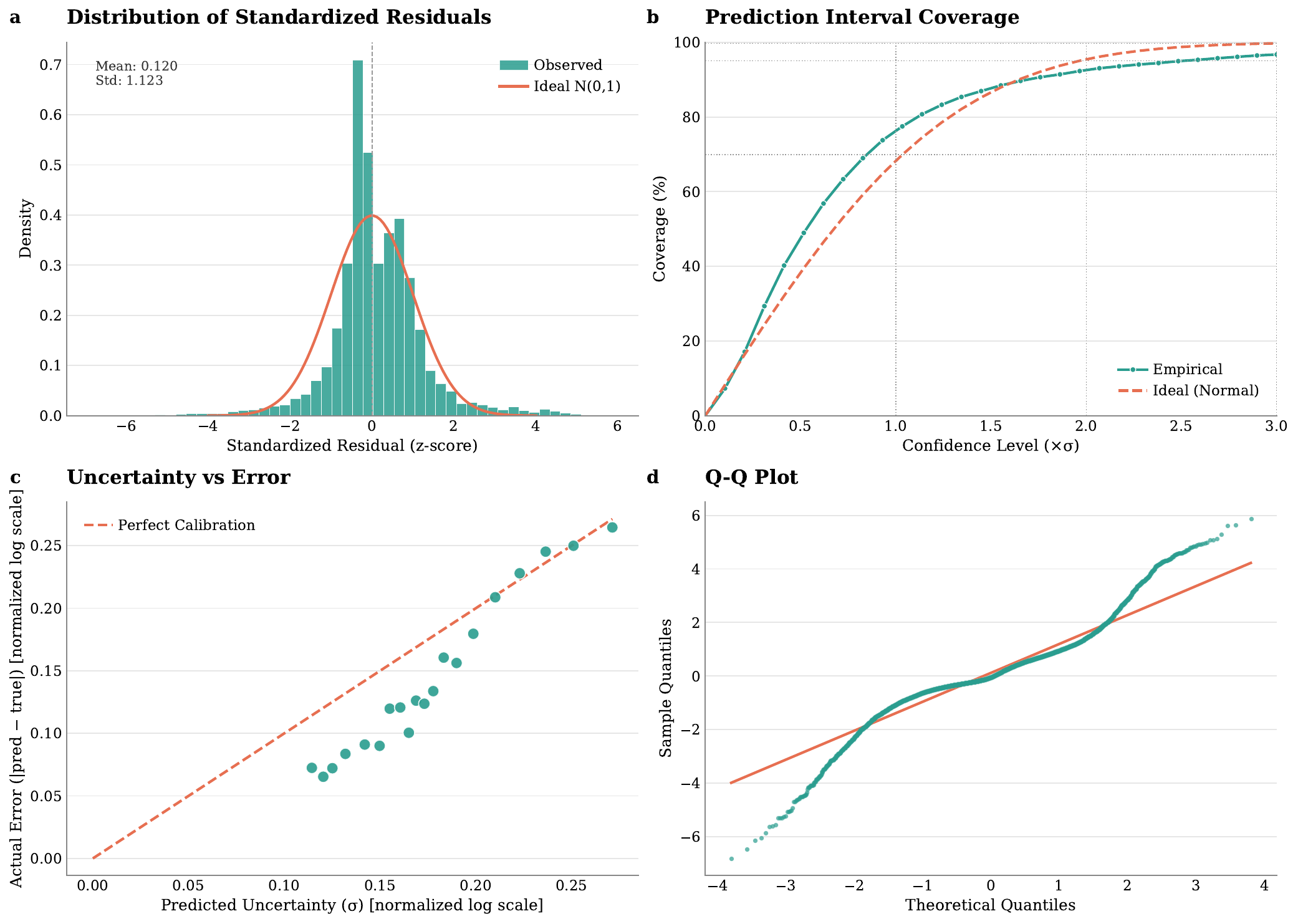}
\caption{Calibration diagnostics for ANP on Guaviare test set (representative run). \textbf{a}: Distribution of standardized residuals closely follows theoretical N(0,1). \textbf{b}: Prediction interval coverage shows slight conservatism. \textbf{c}: Binned uncertainty vs error plot shows positive correlation. Higher predicted uncertainty corresponds to larger actual errors. \textbf{d}: Quantile-quantile plot shows good agreement in central quantiles ($\pm$2$\sigma$), with slight divergence in tails.}
\label{fig:calibration}
\end{figure}

To examine uncertainty quality in greater detail, we analyze calibration diagnostics on the Guaviare test set (Figure~\ref{fig:calibration}). These diagnostics assess whether predicted uncertainties match empirical error distributions at a fine-grained level.

The distribution of standardized residuals $z = (y_{\text{true}} - y_{\text{pred}})/\sigma_{\text{pred}}$ follows the theoretical N(0,1) distribution (Figure~\ref{fig:calibration}, top left). The empirical mean is close to zero, indicating low systemic bias and standard deviation is close to one, indicating well-scaled uncertainties. The histogram aligns well with the theoretical density curve across the full range of residuals, from the left tail ($z < -3$) through the mode to the right tail ($z > 3$). This confirms that ANP's predicted Gaussian distributions accurately characterize the error distribution at individual predictions.

Prediction interval coverage analysis (Figure~\ref{fig:calibration}, top right) reveals slight conservatism. The empirical coverage curve (solid) lies above the theoretical curve (dashed red) for confidence levels up to approximately 1.5$\sigma$, then converges for wider intervals. Specifically, GEDI L4A observations fall within predicted $\pm$1$\sigma$ intervals 83\% of the time, compared to the nominal 68.3\% for Gaussian distributions. This excess indicates the model is cautious and generates prediction intervals that are slightly wider than strictly necessary rather than overconfident. This pattern of conservatism at narrow intervals with convergence at wider intervals is less harmful than undercoverage for applications where false confidence is more costly than slight underconfidence, such as conservation planning under uncertainty.

The relationship between predicted uncertainty and actual error (Figure~\ref{fig:calibration}, bottom left) demonstrates that ANP's uncertainties are well-scaled globally and also informative at individual predictions. We bin predictions by predicted standard deviation and compute mean absolute error within each bin. The resulting curve shows strong positive correlation. Predictions with higher uncertainty exhibit larger actual errors, while predictions with lower uncertainty are more accurate. ANP's curve lies slightly below the diagonal for low-uncertainty predictions (actual errors smaller than predicted) and slightly above for high-uncertainty predictions (actual errors larger than predicted), consistent with the conservative tendency observed in coverage statistics. Importantly, the monotonic increasing relationship confirms that predicted uncertainties provide useful information.

The quantile-quantile plot (Figure~\ref{fig:calibration}, bottom right) compares empirical quantiles of standardized residuals to theoretical quantiles of N(0,1). Points lying on the diagonal indicate perfect agreement. ANP shows nominal calibration in the central quantiles (approximately $-2 < z < 2$, covering ~95\% of predictions), with points closely following the diagonal line. Deviation appears primarily in the tails where the empirical distribution has slightly heavier left tail (more extreme negative residuals) and thinner right tail (fewer extreme positive residuals) than the theoretical Gaussian. The tail deviations are minor and affect only the outermost 2-3\% of predictions, while the bulk of the distribution (97\% of cases) shows strong agreement with the Gaussian assumption.

Together, these diagnostics establish that ANP produces calibrated uncertainties that are globally well-scaled (Z-std $\approx$ 1.0) individually informative (uncertainty correlates with error), and well-approximated by Gaussian distributions except in extreme tails. This level of calibration quality represents an improvement over ensemble tree based methods.

\subsection{Spatial Structure of Uncertainty}

\begin{figure}[H]
\centering
\includegraphics[width=\textwidth]{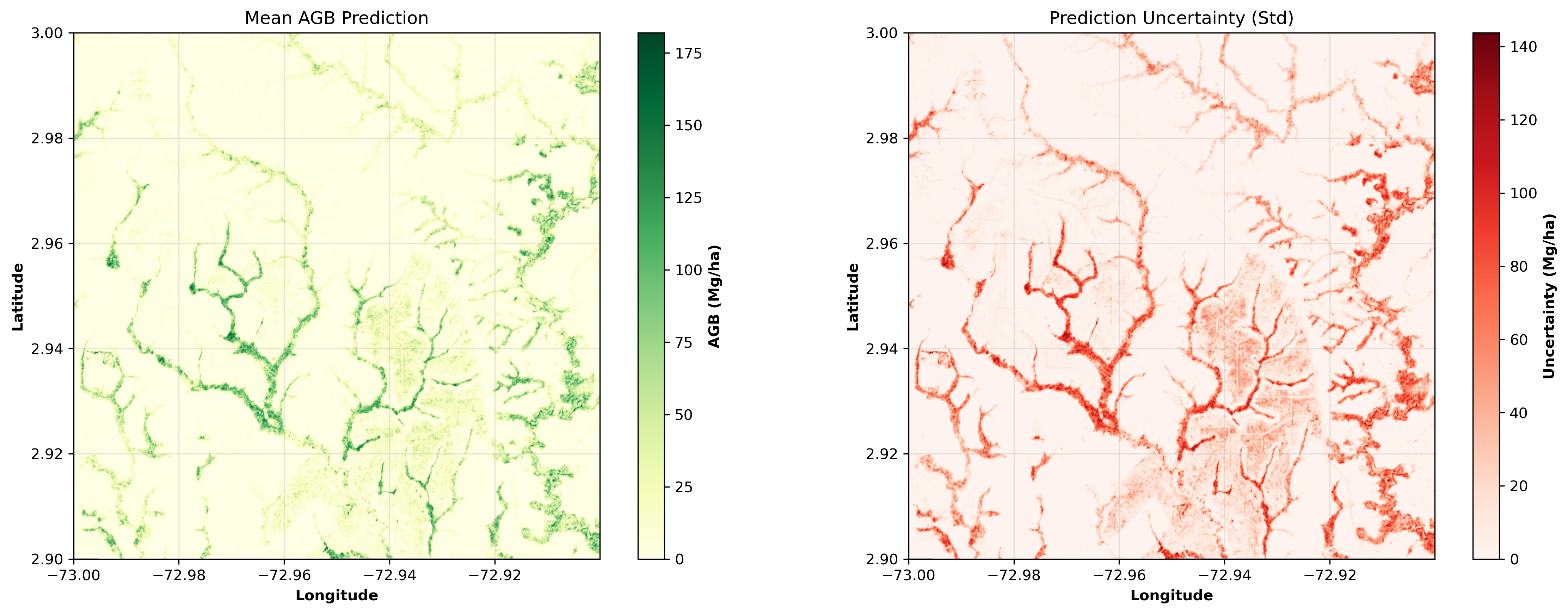}
\caption{Spatial predictions and uncertainty for a 0.1$^\circ \times$ 0.1$^\circ$ tile in Guaviare, Colombia. \textbf{Left}: Mean AGBD prediction shows dendritic patterns of high biomass (green) corresponding to remaining forest corridors, with low biomass in cleared areas (yellow). \textbf{Right}: Predicted standard deviation (uncertainty) is spatially heterogeneous, with highest uncertainty in structurally complex forested areas and lowest uncertainty in homogeneous cleared land. The spatial pattern of uncertainty adapts to land cover configuration, increasing in regions with complex forest structure and decreasing in simple agricultural landscapes.}
\label{fig:spatial_maps}
\end{figure}

Beyond global calibration metrics, we examine whether ANP's uncertainties adapt appropriately to spatial context. Figure~\ref{fig:spatial_maps} shows predictions and uncertainties for a representative 0.1$^\circ \times$ 0.1$^\circ$ tile in Guaviare, Colombia.

The predicted biomass map (Figure~\ref{fig:spatial_maps}, left panel) shows spatial structure. High AGBD values (green, 100-175 Mg/ha) form dendritic branching patterns representing remaining forest corridors, while low values (yellow, 0-50 Mg/ha) indicate cleared agricultural land. These patterns reflect landscape scale deforestation in the region, where forests have been cleared for agriculture and cattle ranching, leaving irregular patches and corridors of remaining vegetation. The fine scale spatial detail (10m resolution) captures individual forest fragments and clearing boundaries, demonstrating the value of foundation model embeddings combined with GEDI's high quality observations.

The uncertainty map (Figure~\ref{fig:spatial_maps}, right panel) shows spatial heterogeneity that corresponds to land cover characteristics. Predicted uncertainty is highest in the densely forested areas showing complex dendritic structure, which are regions where biomass varies substantially over short distances and GEDI footprints may be sparse. Uncertainty is lowest in the cleared agricultural areas. In these homogeneous landscapes, biomass is uniformly low and predictions are straightforward.

\begin{figure}[H]
\centering
\includegraphics[width=\textwidth]{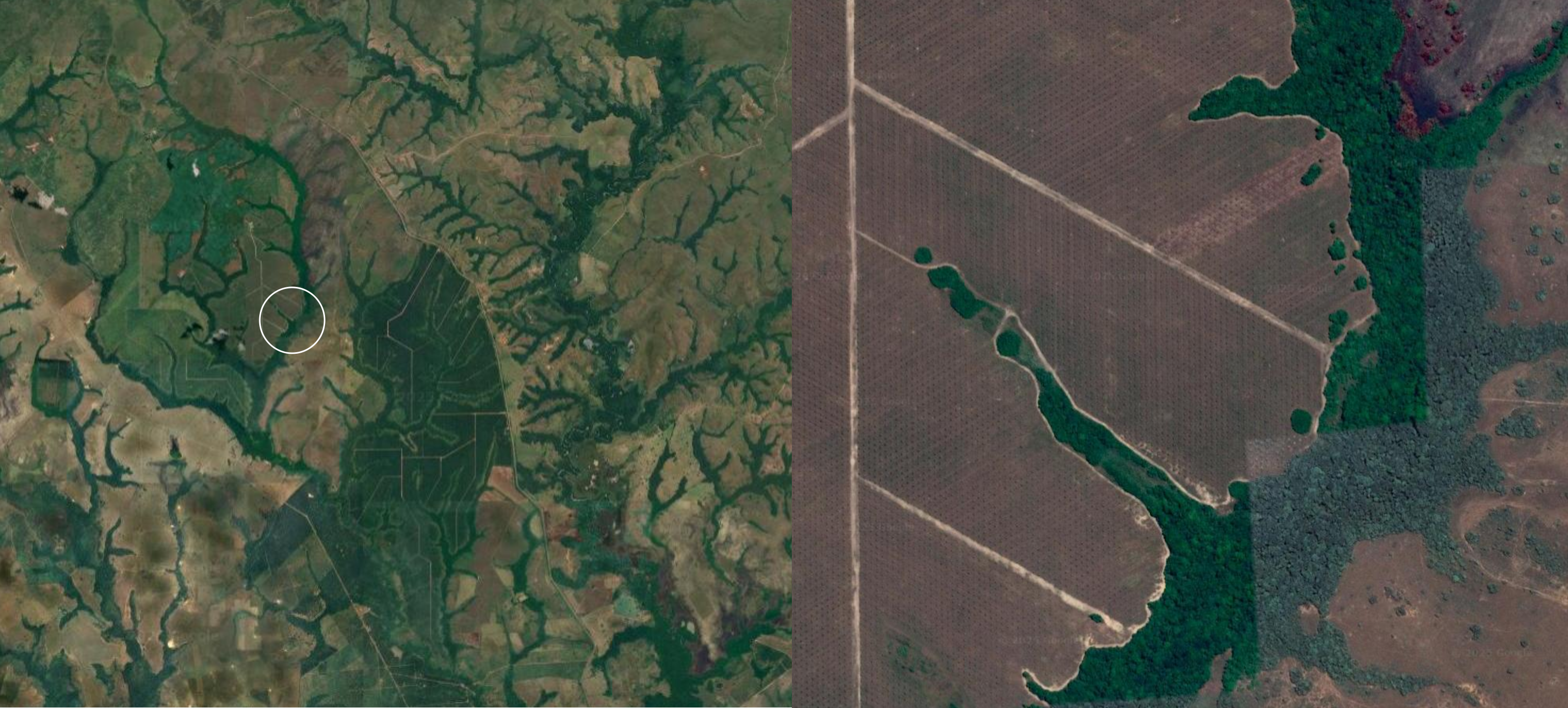}
\caption{Satellite imagery validation of spatial predictions in Guaviare, Colombia. \textbf{Left}: Overview of the prediction area showing landscape scale deforestation patterns. Dark green areas represent intact or planted forest, while tan/brown areas indicate cleared agricultural land. White circle indicates zoom area shown in right panel. \textbf{Right}: Detail of agricultural parcels interspersed with riparian forest corridors. Imagery from Google Maps, map data from Maxar.}
\label{fig:guaviare_satellite}
\end{figure}

Comparison with satellite imagery (Figures~\ref{fig:guaviare_satellite}) shows that uncertainty patterns correspond to actual vegetation structure visible at 10m resolution. The high uncertainty regions align with intact forest patches showing complex canopy structure, tree rows in plantation forestry, and forest-agriculture ecotones where biomass transitions occur over tens of meters. Low uncertainty regions align with large contiguous cleared areas showing uniform spectral characteristics. This qualitative validation indicates that ANP's uncertainty is not arbitrary but rather reflects differences in prediction difficulty driven by landscape heterogeneity.

To quantify the relationship between observation density and predicted uncertainty, we computed block-level correlations for a representative tile. Dividing the tile into $\sim$1~km blocks (0.01$^{\circ}$), we calculated GEDI shot density and mean predicted uncertainty per block. Predicted uncertainty correlates significantly with observation sparsity (Pearson $r = 0.43$, $p < 10^{-5}$). Blocks with fewer GEDI shots show higher predicted uncertainty, confirming that the attention mechanism appropriately expands intervals in data-sparse areas.

\section{Discussion}
\label{discussions}

\subsection{Context-Aware Uncertainty for Heterogeneous Landscapes}
\label{discussions:context}

\begin{figure}[H]
\centering
\hspace*{-2cm}
\includegraphics[width=1.3\textwidth]{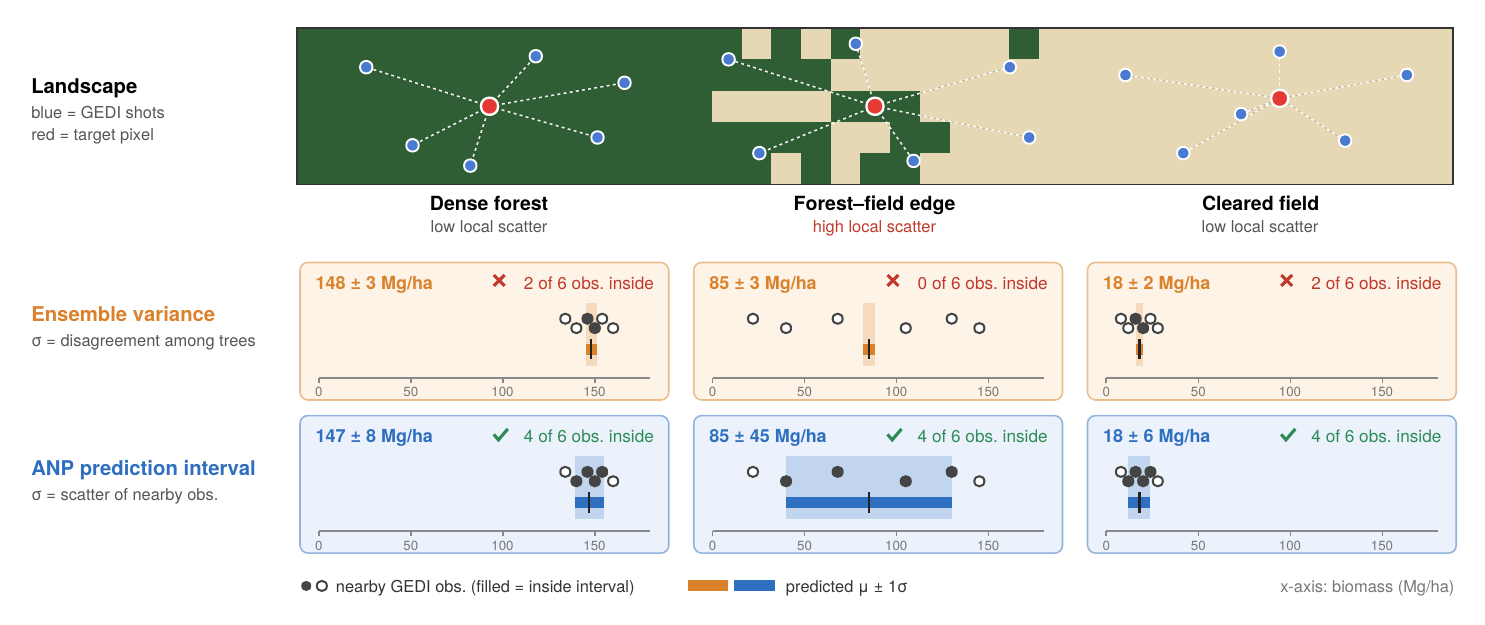}
\caption{Conceptual illustration of ensemble variance vs.\ context-aware prediction intervals at three landscape types. \textit{Top:} each target pixel (red) and its nearby observations (blue). \textit{Rows:} predicted $\mu \pm 1\sigma$ (shaded) against the nearby observations (filled = inside the interval).}
\label{fig:uncertainty_sources}
\end{figure}

Ensemble variance and prediction intervals are often both presented as ``uncertainty,'' but they measure different properties (Figure~\ref{fig:uncertainty_sources}). Ensemble variance quantifies model disagreement across bootstrap samples, while prediction intervals quantify the range within which future observations are likely to fall. Tight ensemble agreement does not imply narrow empirical prediction intervals if the underlying data exhibit high scatter.

Our results demonstrate this discrepancy empirically. Random Forest exhibits overconfident uncertainty estimates from using tree variance agreement as a proxy. XGBoost quantile regression partially addresses this by directly estimating the conditional distribution, but the highest-accuracy configurations remain overconfident, while better-calibrated configurations sacrifice accuracy. The ANP achieves both high accuracy and nearly nominal calibration by conditioning prediction intervals from observation scatter.

Spatially-varying uncertainty is crucial for biomass mapping in heterogeneous landscapes. A cleared field, an intact forest, and a forest-field boundary may have similar spectral features but vastly different biomass variability. Even with embeddings at the target, they learn only global trends (e.g. ``variance is higher near the equator''), but not the scatter of nearby observations.

The ANP's attention mechanism addresses this by conditioning predictions on the set of local context points. The model learns to output wider intervals when context points show high scatter (e.g. at ecotones) and narrower intervals when context points are homogeneous (e.g. in agricultural fields). This adaptation is visible in Figure~\ref{fig:uncertainty_sources}, where uncertainty is highest in complex forest patches and lowest in uniform cleared areas, matching patterns of landscape heterogeneity.

Empirically, predicted uncertainty correlates with GEDI shot density ($r=0.43$), with higher uncertainty in low-density regions. This spatial adaptivity is a key advantage of the ANP over models that treat each prediction independently given the target features. The heteroscedastic output captures aleatoric uncertainty, while the context-conditioning captures the spatial structure of that uncertainty.

The calibration advantage of context-conditioning can be isolated by comparing the ANP to a heteroscedastic deep ensemble trained with the same Gaussian negative log-likelihood loss (\ref{appendix:deep_ensemble}). Both models learn location-dependent variance from features, but only the ANP conditions predictions on the set of nearby GEDI observations via cross-attention. The deep ensemble achieves a better Z-score standard deviation than Random Forest or standard MC Dropout, confirming that the heteroscedastic loss is beneficial, but still overconfident relative to the ANP. This represents the effects of predicting uncertainty from target features alone, without observing the local scatter among context observations.

\subsection{Implications for Operational Mapping}

Miscalibrated uncertainty has consequences for applications. Conservation planning, restoration prioritization, and carbon accounting rely on biomass maps to allocate resources and quantify carbon stocks. Overconfident predictions risk misinformed decisions and inflated carbon credits. Conversely, overly conservative estimates risk undervaluing forest carbon and missing opportunities for intervention.

Our results suggest many published maps may contain unreliable uncertainties. The default RandomForest and XGBoost configurations common in biomass mapping overestimate confidence by 30-45 percentage points at 1$\sigma$. Even with quantile regression, models optimized for accuracy tend to underestimate uncertainty. For downstream uses that depend on reliable intervals, miscalibration at this scale undermines the utility of estimated uncertainties. 

The ANP's context-conditioning provides a mechanism to further adapt intervals to local heterogeneity. For large-scale mapping where both accuracy and calibration are essential, models that combine aleatoric estimation with spatial context awareness offer a promising direction.

\section{Limitations}
\label{limitations}

The ANP's uncertainty estimates arise from high-dimensional feature interactions in the attention mechanism and latent path, which are less transparent than a tree ensemble's decision paths. Feature importance methods for deep learning could partially address this, but a more interesting direction may be to use the ANP as the predictive model while using a simpler surrogate (e.g. linear regression) to explain its predictions. This would decouple the objectives of maximizing accuracy, calibrating uncertainty, and providing interpretable feature attributions.

Our approach treats biomass mapping as a static problem, using a single year of GEDI and Sentinel data. Extending the model to capture temporal dynamics like forest growth, disturbance, and seasonality is nontrivial due to GEDI's uneven temporal coverage and Sentinel's compositing artifacts but could enable monitoring of biomass change. Formalizing this extension, and evaluating whether the ANP's uncertainty estimates remain calibrated under temporal nonstationarity, is a key direction for future work.

By using embeddings as frozen features, our framework treats the spectral-to-embedding mapping as deterministic. Any epistemic uncertainty in the foundation model arising from its pretraining data, architecture choices, or temporal compositing of Sentinel imagery is not accounted for the downstream biomass predictor. Propagating uncertainty through frozen foundation model encoders remains an open problem more broadly, and quantifying how much total predictive uncertainty is attributable to the feature extraction stage versus the biomass regression stage is a direction for future work.

The current framework treats GEDI L4A estimates as point observations. By treating all context observations as equally reliable, the model conflates measurement noise (reducible with better instruments) with spatial heterogeneity (irreducible given the feature set). Propagating GEDI's reported shot-level uncertainties through the model, either via uncertainty-aware attention or by conditioning the latent path on aggregate context uncertainty, could help disentangle these sources and provide more principled estimates when GEDI data quality varies spatially.

\section{Conclusion}
\label{conclusion}

Reliable uncertainty quantification in spatial biomass mapping has been an ongoing challenge. We demonstrate that Attentive Neural Processes produce calibrated prediction intervals that adapt to spatial context.

We summarize our contributions here. ANPs achieve simultaneously excellent accuracy and calibration while Random Forest and XGBoost face limitations between performance and efficiency or between accuracy and calibration. Then, we reveal that widely used baseline configurations produce miscalibrated uncertainties (30-45\% coverage instead of 68\%), with implications beyond biomass. Many published results likely contain unreliable uncertainty estimates. We demonstrate that foundation model embeddings provide effective features for spatial prediction when combined with probabilistic models, offering a practical pathway for leveraging pretrained representations.

Beyond biomass mapping, this work highlights a need for rigorous uncertainty quantification across spatial environmental prediction. Ensemble variance is not a substitute for calibrated prediction intervals. We hope this demonstration will motivate future studies to validate prediction interval coverage on held out data, check whether uncertainty adapts to spatial heterogeneity, and consider probabilistic spatial models that model observation level variability rather than ensemble disagreement.

\section{Data Availability}

GEDI-L4A data is publicly available through \href{https://doi.org/10.3334/ORNLDAAC/2056}{NASA EarthData}, with gediDB \citep{Besnard2025} used here for access and organization. Tessera \citep{feng2025tessera} embeddings are publicly available through the Python package \href{https://github.com/ucam-eo/geotessera}{geotessera}.



\section{Author Contributions}

RY: Conceptualization; Data curation; Formal analysis; Investigation; Methodology; Validation; Visualization; Software; Writing -- original draft; Writing -- review and editing.

SK: Conceptualization; Supervision; Writing -- review and editing.
    
\section{Acknowledgments}

This work was supported by the Taiwan Cambridge Scholarship from the Cambridge Trust and by funding from Dr. Robert Sansom. We thank Prof. Konrad Schindler for his insightful comments.

\bibliographystyle{elsarticle-num-names} 
\bibliography{cas-refs}

@article{Turton2022Improving,
  title        = {Improving Estimates and Change Detection of Forest Above-Ground Biomass Using Statistical Methods},
  author       = {Turton, Amber E. and Augustin, Nicole H. and Mitchard, Edward T.\ A.},
  journal      = {Remote Sensing},
  year         = {2022},
  volume       = {14},
  number       = {19},
  pages        = {4911},
  doi          = {10.3390/rs14194911},
  publisher    = {Multidisciplinary Digital Publishing Institute (MDPI)},
  url          = {https://doi.org/10.3390/rs14194911}
}

@article{Roy2021GEDI,
  title   = {The impact of geolocation uncertainty on GEDI tropical forest canopy height estimation and change monitoring},
  author  = {Roy, David P. and Kashongwe, Herve B. and Armston, John},
  journal = {Science of Remote Sensing},
  volume  = {4},
  pages   = {100024},
  year    = {2021},
  month   = {December},
  doi     = {10.1016/j.srs.2021.100024},
  issn    = {2666-0172},
  url     = {https://doi.org/10.1016/j.srs.2021.100024}
}

@article{ploton2020spatial,
  title={Spatial validation reveals poor predictive performance of large-scale ecological mapping models},
  author={Ploton, Pierre and Mortier, Fr{\'e}d{\'e}ric and R{\'e}jou-M{\'e}chain, Maxime and Barbier, Nicolas and Picard, Nicolas and Rossi, Vivien and Dormann, Carsten and Cornu, Guillaume and Viennois, Ga{\"e}lle and Bayol, Nicolas and Lyapustin, Alexei and Gourlet-Fleury, Sylvie and P{\'e}lissier, Rapha{\"e}l},
  journal={Nature Communications},
  volume={11},
  number={1},
  pages={4540},
  year={2020},
  publisher={Nature Publishing Group},
  doi={10.1038/s41467-020-18321-y}
}

@article{mukhopadhyay2024computation,
  title={Computation of prediction intervals for forest aboveground biomass predictions using generalized linear models in a large-extent boreal forest region},
  author={Mukhopadhyay, Ritwika and Ekstr{\"o}m, Magnus and Lindberg, Eva and Persson, Henrik J and Saarela, Svetlana and Nilsson, Mats},
  journal={Forestry: An International Journal of Forest Research},
  pages={cpae006},
  year={2024},
  publisher={Oxford University Press},
  doi={10.1093/forestry/cpae006}
}

@article{Pascarella2023ReUse,
  title        = {ReUse: REgressive Unet for Carbon Storage and Above-Ground Biomass Estimation},
  author       = {Pascarella, Antonio Elia and Giacco, Giovanni and Rigiroli, Mattia and Marrone, Stefano and Sansone, Carlo},
  journal      = {Journal of Imaging},
  volume       = {9},
  number       = {3},
  pages        = {61},
  year         = {2023},
  doi          = {10.3390/jimaging9030061},
  url          = {https://www.mdpi.com/2313-433X/9/3/61}
}

@article{Araza2022ACF,
  title        = {A comprehensive framework for assessing the accuracy and uncertainty of global above-ground biomass maps},
  author       = {Araza, Arnan B. and de Bruin, Sytze and Herold, Martin and Quegan, Shaun and Labri{\`e}re, Nicolas and Rodr{\'i}guez-Veiga, Pedro and Avitabile, Valerio and Santoro, Maurizio and Mitchard, Edward T. A. and Ryan, Casey M. and Phillips, Oliver L. and Willcock, Simon and Verbeeck, Hans and Carreiras, Jo{\~a}o M. B. and Hein, Lars and Schelhaas, Mart-Jan and Pacheco-Pascagaza, Ana Maria and Bispo, Polyanna da Concei{\c{c}}{\~a}o and Vaglio Laurin, Gaia and Vieilledent, Ghislain and Slik, Ferry J. W. and Wijaya, Arief and Lewis, Simon L. and Morel, Alexandra C. and Liang, Jingjing and Sukhdeo, Hansrajie and Schepaschenko, Dmitry and {\v{C}}avlovi{\'c}, Jura and Gilani, Hammad and Lucas, Richard M.},
  journal      = {Remote Sensing of Environment},
  volume       = {272},
  pages        = {112917},
  year         = {2022},
  doi          = {10.1016/j.rse.2022.112917},
  url          = {https://doi.org/10.1016/j.rse.2022.112917}
}

@article{Doob1935Limiting,
  author    = {Doob, J. L.},
  title     = {The Limiting Distributions of Certain Statistics},
  journal   = {Annals of Mathematical Statistics},
  volume    = {6},
  number    = {3},
  pages     = {160--169},
  year      = {1935},
  month     = sep,
  doi       = {10.1214/aoms/1177732594}
}

@article{Santoro2024DesignAP,
  title={Design and performance of the Climate Change Initiative Biomass global retrieval algorithm},
  author={Maurizio Santoro and Oliver Cartus and Shaun Quegan and Heather Kay and Richard M. Lucas and Arnan B. Araza and Martin Herold and Nicolas Labrie{\`e}re and J{\'e}r{\^o}me Chave and Ake Rosenqvist and Takeo Tadono and Kazufumi Kobayashi and Josef Kellndorfer and Valerio Avitabile and Hugh Brown and Jo{\~a}o Carreiras and Michael J. Campbell and Jura {\v{C}}avlovi{\'c} and Polyanna da Concei{\c{c}}{\~a}o Bispo and Hammad Gilani and Mohammed Latif Khan and Amit Kumar and Simon L. Lewis and Jingjing Liang and Edward T. A. Mitchard and Ana Maria Pacheco Pascagaza and Oliver L. Phillips and Casey M. Ryan and Purabi Saikia and Dmitry Schepaschenko and Hansrajie Sukhdeo and Hans Verbeeck and Ghislain Vieilledent and Arief Wijaya and Simon Willcock and Frank Martin Seifert},
  journal={Science of Remote Sensing},
  year={2024},
  volume={10},
  pages={100169},
  doi={10.1016/j.srs.2024.100169}
}

@article{Duncanson2022GEDI_AGB,
  title = {Aboveground biomass density models for {NASA}’s Global Ecosystem Dynamics Investigation ({GEDI}) lidar mission},
  author = {Duncanson, Laura and Kellner, James R. and Armston, John and Dubayah, Ralph and Minor, David M. and Hancock, Steven and Healey, Sean P. and Patterson, Paul L. and Saarela, Svetlana and Marselis, Suzanne and Silva, Carlos E. and Bruening, Jamis and Goetz, Scott J. and Tang, Hao and Hofton, Michelle and Blair, Bryan and Luthcke, Scott and Fatoyinbo, Lola and Abernethy, Katharine and Alonso, Alfonso and Andersen, Hans-Erik and Aplin, Paul and Baker, Timothy R. and Barbier, Nicolas and Bastin, Jean Francois and Biber, Peter and Boeckx, Pascal and Bogaert, Jan and Boschetti, Luigi and Boucher, Peter Brehm and Boyd, Doreen S. and Burslem, David F.R.P. and Calvo-Rodriguez, Sofia and Chave, Jérôme and Chazdon, Robin L. and Clark, David B. and Clark, Deborah A. and Cohen, Warren B. and Coomes, David A. and Corona, Piermaria and Cushman, K.C. and Cutler, Mark E.J. and Dalling, James W. and Dalponte, Michele and Dash, Jonathan and de-Miguel, Sergio and Deng, Songqiu and Ellis, Peter Woods and Erasmus, Barend and Fekety, Patrick A. and Fernandez-Landa, Alfredo and Ferraz, Antonio and Fischer, Rico and Fisher, Adrian G. and García-Abril, Antonio and Gobakken, Terje and Hacker, Jorg M. and Heurich, Marco and Hill, Ross A. and Hopkinson, Chris and Huang, Huabing and Hubbell, Stephen P. and Hudak, Andrew T. and Huth, Andreas and Imbach, Benedikt and Jeffery, Kathryn J. and Katoh, Masato and Kearsley, Elizabeth and Kenfack, David and Kljun, Natascha and Knapp, Nikolai and Král, Kamil and Krůček, Martin and Labrière, Nicolas and Lewis, Simon L. and Longo, Marcos and Lucas, Richard M. and Main, Russell and Manzanera, Jose A. and Martínez, Rodolfo Vásquez and Mathieu, Renaud and Memiaghe, Herve and Meyer, Victoria and Monteagudo Mendoza, Abel and Monerris, Alessandra and Montesano, Paul and Morsdorf, Felix and Næsset, Erik and Naidoo, Laven and Nilus, Reuben and O’Brien, Michael and Orwig, David A. and Papathanassiou, Konstantinos and Parker, Geoffrey and Philipson, Christopher and Phillips, Oliver L. and Pisek, Jan and Poulsen, John R. and Pretzsch, Hans and Rüdiger, Christoph and Saatchi, Sassan and Sanchez-Azofeifa, Arturo and Sanchez-Lopez, Nuria and Scholes, Robert and Silva, Carlos A. and Simard, Marc and Skidmore, Andrew and Stereńczak, Krzysztof and Tanase, Mihai and Torresan, Chiara and Valbuena, Ruben and Verbeeck, Hans and Vrška, Tomas and Wessels, Konrad and White, Joanne C. and White, Lee J.T. and Zahabu, Eliakimu and Zgraggen, Carlo},
  journal = {Remote Sensing of Environment},
  volume = {270},
  pages = {112845},
  year = {2022},
  month = mar,
  doi = {10.1016/j.rse.2021.112845}
}

@article{Chave2014,
  author    = {Jérôme Chave and Maxime Réjou-Méchain and Alberto Búrquez and Emmanuel Chidumayo and Matthew S. Colgan and Welington B.C. Delitti and Alvaro Duque and Tron Eid and Philip M. Fearnside and Rosa C. Goodman and Matieu Henry and Angelina Martínez-Yrízar and Wilson A. Mugasha and Helene C. Muller-Landau and Maurizio Mencuccini and Bruce W. Nelson and Alfred Ngomanda and Euler M. Nogueira and Edgar Ortiz-Malavassi and Raphaël Pélissier and Pierre Ploton and Casey M. Ryan and Juan G. Saldarriaga and Ghislain Vieilledent},
  title     = {Improved allometric models to estimate the aboveground biomass of tropical trees},
  journal   = {Global Change Biology},
  year      = {2014},
  volume    = {20},
  pages     = {3177--3190},
  doi       = {10.1111/gcb.12629}
}

@article{Song2023,
  author    = {C. M. Song and Z. Xiong and X. X. Zhu},
  title     = {Biomass Estimation and Uncertainty Quantification From Tree Height},
  journal   = {IEEE Journal of Selected Topics in Applied Earth Observations and Remote Sensing},
  volume    = {16},
  pages     = {4833--4845},
  year      = {2023},
  doi       = {10.1109/JSTARS.2023.3271186},
}

@article{koenker1978regression,
  title        = {Regression Quantiles},
  author       = {Koenker, Roger and Bassett, Gilbert Jr.},
  journal      = {Econometrica},
  volume       = {46},
  number       = {1},
  pages        = {33--50},
  year         = {1978},
  publisher    = {The Econometric Society},
  doi          = {10.2307/1913643},
  url          = {https://www.jstor.org/stable/1913643}
}

@misc{iso14064-1-2018,
  title        = {ISO 14064-1:2018 — Greenhouse gases — Part 1: Specification with guidance at the organization level for quantification and reporting of greenhouse gas emissions and removals},
  author = {{International Organization for Standardization}},
  year         = {2018},
  number       = {ISO 14064-1:2018},
  edition      = {2},
  month        = {Dec},
  url          = {https://www.iso.org/standard/66453.html}
}

@InProceedings{Gal2016mcdropout,
  title = 	 {Dropout as a Bayesian Approximation: Representing Model Uncertainty in Deep Learning},
  author = 	 {Gal, Yarin and Ghahramani, Zoubin},
  booktitle = 	 {Proceedings of The 33rd International Conference on Machine Learning},
  pages = 	 {1050--1059},
  year = 	 {2016},
  editor = 	 {Balcan, Maria Florina and Weinberger, Kilian Q.},
  volume = 	 {48},
  series = 	 {Proceedings of Machine Learning Research},
  address = 	 {New York, New York, USA},
  month = 	 {20--22 Jun},
  publisher =    {PMLR},
  url = 	 {https://proceedings.mlr.press/v48/gal16.html},
}

@article{Lang2022GlobalCanopyHeight,
  title   = {Global canopy height regression and uncertainty estimation from GEDI LiDAR waveforms with deep ensembles},
  author  = {Lang, Nico and Kalischek, Nikolai and Armston, John and Schindler, Konrad and Dubayah, Ralph and Wegner, Jan Dirk},
  journal = {Remote Sensing of Environment},
  volume  = {268},
  pages   = {112760},
  year    = {2022},
  month   = {January},
  doi     = {10.1016/j.rse.2021.112760},
  issn    = {0034-4257}
}

@article{Baskerville1972,
  author    = {G. L. Baskerville},
  title     = {Use of Logarithmic Regression in the Estimation of Plant Biomass},
  journal   = {Canadian Journal of Forest Research},
  year      = {1972},
  volume    = {2},
  number    = {1},
  pages     = {49--53},
  doi       = {10.1139/x72-009}
}

@article{Xu2024ForestAB,
  title={Forest aboveground biomass estimation based on spaceborne LiDAR combining machine learning model and geostatistical method},
  author={Li Xu and Jinge Yu and Qingtai Shu and Shaolong Luo and Wenwu Zhou and Dandan Duan},
  journal={Frontiers in Plant Science},
  year={2024},
  volume={15},
  pages={1428268},
  doi={10.3389/fpls.2024.1428268}
}

@article{sialelli2025agbd,
  title = {AGBD: A Global-scale Biomass Dataset},
  author = {Sialelli, Ghjulia and Peters, Torben and Wegner, Jan D. and Schindler, Konrad},
  journal = {ISPRS Annals of the Photogrammetry, Remote Sensing and Spatial Information Sciences},
  volume = {X-G-2025},
  pages = {829--838},
  year = {2025},
  doi = {10.5194/isprs-annals-X-G-2025-829-2025},
  url = {https://doi.org/10.5194/isprs-annals-X-G-2025-829-2025}
}

@article{Dubayah2022GEDILA,
  title={GEDI launches a new era of biomass inference from space},
  author={Ralph O. Dubayah and John David Armston and Sean P. Healey and Jamis M. Bruening and Paul L. Patterson and James R. Kellner and Laura Innice Duncanson and Svetlana Saarela and G{\"o}ran St{\aa}hl and Zhiqiang Yang and Hao Tang and J. Bryan Blair and Lola Fatoyinbo and S. Goetz and Steven Hancock and Matthew K. Hansen and Michelle A. Hofton and George C. Hurtt and Scott B. Luthcke},
  journal={Environmental Research Letters},
  year={2022},
  volume={17},
  number={9},
  pages={095001},
  doi={10.1088/1748-9326/ac8694}
}

@article{Zurqani2025AMA,
  title={A multi-source approach combining GEDI LiDAR, satellite data, and machine learning algorithms for estimating forest aboveground biomass on Google Earth Engine platform},
  author={Hamdi A. Zurqani},
  journal={Ecological Informatics},
  year={2025},
  volume={86},
  pages={103052},
  doi={10.1016/j.ecoinf.2025.103052}
}

@article{Li2020,
  author       = {Yingchang Li and Mingyang Li and Chao Li and Zhenzhen Liu},
  title        = {Forest aboveground biomass estimation using Landsat 8 and Sentinel-1A data with machine learning algorithms},
  journal      = {Scientific Reports},
  volume       = {10},
  pages        = {9952},
  year         = {2020},
  doi          = {10.1038/s41598-020-67024-3},
  url          = {https://doi.org/10.1038/s41598-020-67024-3}
}

@article{Dubayah2020GEDI,
  title   = {The Global Ecosystem Dynamics Investigation: High-resolution laser ranging of the Earth’s forests and topography},
  author  = {Dubayah, Ralph and Blair, James Bryan and Goetz, Scott and Fatoyinbo, Lola and Hansen, Matthew and Healey, Sean and Hofton, Michelle and Hurtt, George and Kellner, James and Luthcke, Scott and Armston, John and Tang, Hao and Duncanson, Laura and Hancock, Steven and Jantz, Patrick and Marselis, Suzanne and Patterson, Paul L. and Qi, Wenlu and Silva, Carlos},
  journal = {Science of Remote Sensing},
  volume  = {1},
  pages   = {100002},
  year    = {2020},
  month   = {June},
  doi     = {10.1016/j.srs.2020.100002},
  url     = {https://doi.org/10.1016/j.srs.2020.100002}
}

@article{Besnard2025, doi = {10.21105/joss.08593}, url = {https://doi.org/10.21105/joss.08593}, year = {2025}, publisher = {The Open Journal}, volume = {10}, number = {113}, pages = {8593}, author = {Besnard, Simon and Dombrowski, Felix and Holcomb, Amelia}, title = {gediDB: A toolbox for processing and providing Global Ecosystem Dynamics Investigation (GEDI) L2A-B and L4A-C data}, journal = {Journal of Open Source Software} }

@InProceedings{feng2025tessera,
    author    = {Feng, Zhengpeng and Atzberger, Clement and Jaffer, Sadiq and Knezevic, Jovana and Sormunen, Silja and Young, Robin and Lisaius, Madeline C. and Immitzer, Markus and Jackson, Toby and Ball, James and Coomes, David A. and Madhavapeddy, Anil and Blake, Andrew and Keshav, Srinivasan},
    title     = {TESSERA: Temporal Embeddings of Surface Spectra for Earth Representation and Analysis},
    booktitle = {Proceedings of the IEEE/CVF Conference on Computer Vision and Pattern Recognition (CVPR)},
    month     = {June},
    year      = {2026},
    pages     = {34818-34831},
    url={https://openaccess.thecvf.com/content/CVPR2026/papers/Feng_TESSERA_Temporal_Embeddings_of_Surface_Spectra_for_Earth_Representation_and_CVPR_2026_paper.pdf},
}

@article{Shendryk2022_FusingGEDI,
  title        = {Fusing GEDI with Earth Observation Data for Large Area Aboveground Biomass Mapping},
  author       = {Shendryk, Yuri},
  journal      = {International Journal of Applied Earth Observation and Geoinformation},
  volume       = {115},
  pages        = {103108},
  year         = {2022},
  publisher    = {Elsevier},
  doi          = {10.1016/j.jag.2022.103108},
  url          = {https://doi.org/10.1016/j.jag.2022.103108}
}

@inproceedings{kim2018attentive,
title={Attentive Neural Processes},
author={Hyunjik Kim and Andriy Mnih and Jonathan Schwarz and Marta Garnelo and Ali Eslami and Dan Rosenbaum and Oriol Vinyals and Yee Whye Teh},
booktitle={International Conference on Learning Representations},
year={2019},
url={https://openreview.net/forum?id=SkE6PjC9KX},
}

@InProceedings{garnelo2018conditional,
  title = 	 {Conditional Neural Processes},
  author =       {Garnelo, Marta and Rosenbaum, Dan and Maddison, Christopher and Ramalho, Tiago and Saxton, David and Shanahan, Murray and Teh, Yee Whye and Rezende, Danilo and Eslami, S. M. Ali},
  booktitle = 	 {Proceedings of the 35th International Conference on Machine Learning},
  pages = 	 {1704--1713},
  year = 	 {2018},
  editor = 	 {Dy, Jennifer and Krause, Andreas},
  volume = 	 {80},
  series = 	 {Proceedings of Machine Learning Research},
  month = 	 {10--15 Jul},
  publisher =    {PMLR},
  url = 	 {https://proceedings.mlr.press/v80/garnelo18a.html},
}

@inproceedings{nascetti2023biomassters,
author = {Nascetti, Andrea and Yadav, Ritu and Brodt, Kiril and Qu, Qixun and Fan, Hongwei and Shendryk, Yuri and Shah, Isha and Chung, Christine},
title = {BioMassters: a benchmark dataset for forest biomass estimation using multi-modal satellite time series},
year = {2023},
publisher = {Curran Associates Inc.},
address = {Red Hook, NY, USA},
booktitle = {Proceedings of the 37th International Conference on Neural Information Processing Systems},
articleno = {896},
numpages = {12},
location = {New Orleans, LA, USA},
series = {NIPS '23},
url = {https://openreview.net/pdf?id=hrWsIC4Cmz}
}

@article{Mentch2016rf,
  author  = {Lucas Mentch and Giles Hooker},
  title   = {Quantifying Uncertainty in Random Forests via Confidence Intervals and Hypothesis Tests},
  journal = {Journal of Machine Learning Research},
  year    = {2016},
  volume  = {17},
  number  = {26},
  pages   = {1--41},
  url     = {http://jmlr.org/papers/v17/14-168.html}
}

@article{wager2018estimation,
  title={Estimation and Inference of Heterogeneous Treatment Effects Using Random Forests},
  author={Wager, Stefan and Athey, Susan},
  journal={Journal of the American Statistical Association},
  volume={113},
  number={523},
  pages={1228--1242},
  year={2018},
  doi={10.1080/01621459.2017.1319839}
}

@inproceedings{Lakshminarayanan2017scalable,
author = {Lakshminarayanan, Balaji and Pritzel, Alexander and Blundell, Charles},
title = {Simple and scalable predictive uncertainty estimation using deep ensembles},
year = {2017},
isbn = {9781510860964},
publisher = {Curran Associates Inc.},
address = {Red Hook, NY, USA},
booktitle = {Proceedings of the 31st International Conference on Neural Information Processing Systems},
pages = {6405–6416},
numpages = {12},
location = {Long Beach, California, USA},
series = {NIPS'17},
url = {https://arxiv.org/abs/1612.01474}
}

@InProceedings{Kuleshov2018accurate,
  title = 	 {Accurate Uncertainties for Deep Learning Using Calibrated Regression},
  author =       {Kuleshov, Volodymyr and Fenner, Nathan and Ermon, Stefano},
  booktitle = 	 {Proceedings of the 35th International Conference on Machine Learning},
  pages = 	 {2796--2804},
  year = 	 {2018},
  editor = 	 {Dy, Jennifer and Krause, Andreas},
  volume = 	 {80},
  series = 	 {Proceedings of Machine Learning Research},
  month = 	 {10--15 Jul},
  publisher =    {PMLR},
  url = 	 {https://proceedings.mlr.press/v80/kuleshov18a.html},
}

@book{Rasmussen2005GP,
    author = {Rasmussen, Carl Edward and Williams, Christopher K. I.},
    title = {Gaussian Processes for Machine Learning},
    publisher = {The MIT Press},
    year = {2005},
    month = {11},
    isbn = {9780262256834},
    doi = {10.7551/mitpress/3206.001.0001},
    url = {https://doi.org/10.7551/mitpress/3206.001.0001}
}

@inproceedings{Chen2016xgboost,
author = {Chen, Tianqi and Guestrin, Carlos},
title = {XGBoost: A Scalable Tree Boosting System},
year = {2016},
isbn = {9781450342322},
publisher = {Association for Computing Machinery},
address = {New York, NY, USA},
url = {https://doi.org/10.1145/2939672.2939785},
doi = {10.1145/2939672.2939785},
booktitle = {Proceedings of the 22nd ACM SIGKDD International Conference on Knowledge Discovery and Data Mining},
pages = {785–794},
numpages = {10},
location = {San Francisco, California, USA},
series = {KDD '16}
}

@article{breiman2001random,
  author    = {Breiman, Leo},
  title     = {Random Forests},
  journal   = {Machine Learning},
  volume    = {45},
  number    = {1},
  pages     = {5--32},
  year      = {2001},
  doi       = {10.1023/A:1010933404324},
  url       = {https://doi.org/10.1023/A:1010933404324}
}

@article{Duncanson2023ProtectedAreas,
  author    = {Duncanson, L. and Liang, M. and Leitold, V. and Armston, J. and Krishna Moorthy, S. M. and Dubayah, R. and Costedoat, S. and Enquist, B. J. and Fatoyinbo, L. and Goetz, S. J. and Gonzalez-Roglich, M. and Merow, C. and Roehrdanz, P. R. and Tabor, K. and Zvoleff, A.},
  title     = {The effectiveness of global protected areas for climate change mitigation},
  journal   = {Nature Communications},
  year      = {2023},
  volume    = {14},
  number    = {1},
  pages     = {2908},
  doi       = {10.1038/s41467-023-38073-9},
  url       = {https://doi.org/10.1038/s41467-023-38073-9},
  publisher = {Nature Publishing Group}
}

@inproceedings{higgins2017betavae,
title={beta-{VAE}: Learning Basic Visual Concepts with a Constrained Variational Framework},
author={Irina Higgins and Loic Matthey and Arka Pal and Christopher Burgess and Xavier Glorot and Matthew Botvinick and Shakir Mohamed and Alexander Lerchner},
booktitle={International Conference on Learning Representations},
year={2017},
url={https://openreview.net/forum?id=Sy2fzU9gl}
}

@inproceedings{loshchilov2018decoupled,
title={Decoupled Weight Decay Regularization},
author={Ilya Loshchilov and Frank Hutter},
booktitle={International Conference on Learning Representations},
year={2019},
url={https://openreview.net/forum?id=Bkg6RiCqY7},
}

@article{Roberts2017,
  author    = {Roberts, D. R. and Bahn, V. and Ciuti, S. and Boyce, M. S. and Elith, J. and Guillera-Arroita, G. and Hauenstein, S. and Lahoz-Monfort, J. J. and Schr\"oder, B. and Thuiller, W. and Warton, D. I. and Wintle, B. A. and Hartig, F. and Dormann, C. F.},
  title     = {Cross-validation strategies for data with temporal, spatial, hierarchical, or phylogenetic structure},
  journal   = {Ecography},
  year      = {2017},
  volume    = {40},
  pages     = {913--929},
  doi       = {10.1111/ecog.02881}
}

@article{Carreiras2017Coverage,
  title        = {Coverage of high biomass forests by the ESA BIOMASS mission under defense restrictions},
  author       = {Carreiras, Jo{\~a}o M. B. and Quegan, Shaun and Le Toan, Thuy and Minh, Dinh Ho Tong and Saatchi, Sassan S. and Carvalhais, Nuno and Reichstein, Markus and Scipal, Klaus},
  journal      = {Remote Sensing of Environment},
  volume       = {196},
  pages        = {154--162},
  year         = {2017},
  doi          = {10.1016/j.rse.2017.05.003},
  publisher    = {Elsevier}
}

@misc{zhu2024foundationsearthclimatefoundation,
    title={On the Foundations of Earth and Climate Foundation Models}, 
    author={Xiao Xiang Zhu and Zhitong Xiong and Yi Wang and Adam J. Stewart and Konrad Heidler and Yuanyuan Wang and Zhenghang Yuan and Thomas Dujardin and Qingsong Xu and Yilei Shi},
    journal = {Communications Earth \& Environment},
    volume  = {7},
    pages   = {103},
    year    = {2026},
    doi     = {10.1038/s43247-025-03127-x},
    url     = {https://doi.org/10.1038/s43247-025-03127-x}
}

@misc{huang2025surveyremotesensingfoundation,
      title={A Survey on Remote Sensing Foundation Models: From Vision to Multimodality}, 
      author={Ziyue Huang and Hongxi Yan and Qiqi Zhan and Shuai Yang and Mingming Zhang and Chenkai Zhang and YiMing Lei and Zeming Liu and Qingjie Liu and Yunhong Wang},
      year={2025},
      eprint={2503.22081},
      archivePrefix={arXiv},
      primaryClass={cs.CV},
      url={https://arxiv.org/abs/2503.22081}, 
}

@article{OtavianoDeAlmeida2025,
  author       = {Otaviano, J. C. R. and de Almeida, C. F. P.},
  title        = {Accessing the spatial distribution of aboveground biomass in tropical mountain forests using regression kriging simulation: a geostatistical approach for local‑scale estimates},
  journal      = {Ecological Processes},
  year         = {2025},
  volume       = {14},
  number       = {44},
  doi          = {10.1186/s13717-025-00590-4},
  url          = {https://doi.org/10.1186/s13717-025-00590-4},
  issn         = {2097-1311},
}

@Article{RejouMechain2014,
AUTHOR = {R\'ejou-M\'echain, M. and Muller-Landau, H. C. and Detto, M. and Thomas, S. C. and Le Toan, T. and Saatchi, S. S. and Barreto-Silva, J. S. and Bourg, N. A. and Bunyavejchewin, S. and Butt, N. and Brockelman, W. Y. and Cao, M. and C\'ardenas, D. and Chiang, J.-M. and Chuyong, G. B. and Clay, K. and Condit, R. and Dattaraja, H. S. and Davies, S. J. and Duque, A. and Esufali, S. and Ewango, C. and Fernando, R. H. S. and Fletcher, C. D. and Gunatilleke, I. A. U. N. and Hao, Z. and Harms, K. E. and Hart, T. B. and H\'erault, B. and Howe, R. W. and Hubbell, S. P. and Johnson, D. J. and Kenfack, D. and Larson, A. J. and Lin, L. and Lin, Y. and Lutz, J. A. and Makana, J.-R. and Malhi, Y. and Marthews, T. R. and McEwan, R. W. and McMahon, S. M. and McShea, W. J. and Muscarella, R. and Nathalang, A. and Noor, N. S. M. and Nytch, C. J. and Oliveira, A. A. and Phillips, R. P. and Pongpattananurak, N. and Punchi-Manage, R. and Salim, R. and Schurman, J. and Sukumar, R. and Suresh, H. S. and Suwanvecho, U. and Thomas, D. W. and Thompson, J. and Ur\'{\i}arte, M. and Valencia, R. and Vicentini, A. and Wolf, A. T. and Yap, S. and Yuan, Z. and Zartman, C. E. and Zimmerman, J. K. and Chave, J.},
TITLE = {Local spatial structure of forest biomass and its consequences for remote sensing of carbon stocks},
JOURNAL = {Biogeosciences},
VOLUME = {11},
YEAR = {2014},
NUMBER = {23},
PAGES = {6827--6840},
URL = {https://bg.copernicus.org/articles/11/6827/2014/},
DOI = {10.5194/bg-11-6827-2014}
}

@inproceedings{kingma2014vae,
  author    = {Kingma, Diederik P. and Welling, Max},
  title     = {Auto-Encoding Variational Bayes},
  booktitle = {Proceedings of the 2nd International Conference on Learning Representations (ICLR 2014)},
  year      = {2014},
  pages     = {1--14},
  publisher = {arXiv},
  address   = {Ithaca, NY},
  eprint    = {1312.6114},
  archivePrefix = {arXiv},
  url       = {https://arxiv.org/abs/1312.6114}
}

@article{Li2022RF_MC,
  author    = {Li Zizhao and Bi Shoudong and Hao Shuang and Cui Yuhuan},
  title     = {Aboveground biomass estimation in forests with random forest and {Monte Carlo}-based uncertainty analysis},
  journal   = {Ecological Indicators},
  volume    = {142},
  pages     = {109246},
  year      = {2022},
  doi       = {10.1016/j.ecolind.2022.109246}
}

@article{Yang2023China,
  author = {Yang Qiuli and Chunyue Niu and Xiaoqiang Liu and Yuhao Feng and Qin Ma and Xuejing Wang and Hao Tang and Qinghua Guo},
  title     = {Mapping high-resolution forest aboveground biomass of {China} using multisource remote sensing data},
  journal   = {GIScience \& Remote Sensing},
  volume    = {60},
  number    = {1},
  pages     = {2203303},
  year      = {2023},
  doi       = {10.1080/15481603.2023.2203303}
}

@article{Liu2024Shrub,
  author = {Wenchao Liu and Jie Wang and Yang Hu and Taiyong Ma and Munkhdulam Otgonbayar and Chunbo Li and You Li and Jilin Yang},
  title     = {Mapping Shrub Biomass at 10\,m Resolution by Integrating Field Measurements, Unmanned Aerial Vehicles, and Multi-Source Satellite Observations},
  journal   = {Remote Sensing},
  volume    = {16},
  number    = {16},
  pages     = {3095},
  year      = {2024},
  doi       = {10.3390/rs16163095}
}

@article{May2024_Indonesia_AGBD,
  title   = {Mapping aboveground biomass in Indonesian lowland forests using GEDI and hierarchical models},
  author  = {May, Paul B. and Schlund, Michael and Armston, John D. and Kotowska, Martyna M. and Brambach, Fabian and Wenzel, Arne and Erasmi, Stefan},
  journal = {Remote Sensing of Environment},
  volume  = {313},
  pages   = {114384},
  year    = {2024},
  month   = {November},
  doi     = {10.1016/j.rse.2024.114384},
  url     = {https://doi.org/10.1016/j.rse.2024.114384}
}

\appendix

\section{Neural Process Architecture and Training Objective}
\label{app:architecture}

Figure~\ref{fig:anp_architecture} provides an architectural diagram overview of the ANP architecture. The following sections specify each component formally.

\begin{figure}[htbp]
\centering
\resizebox{\textwidth}{!}{\begin{tikzpicture}[
    font=\small,
    >={Stealth[length=2.5mm]},
    smallbox/.style={
        draw, rounded corners=2pt, minimum height=6mm, minimum width=20mm,
        align=center, fill=white, font=\footnotesize
    },
    input/.style={
        draw, rounded corners=2pt, minimum height=8mm, minimum width=30mm,
        align=center, fill=blue!8, thick
    },
    encoder/.style={
        draw, rounded corners=2pt, minimum height=9mm, minimum width=34mm,
        align=center, fill=orange!15, thick
    },
    detbox/.style={
        draw, rounded corners=2pt, minimum height=9mm, minimum width=40mm,
        align=center, fill=green!12, thick
    },
    latbox/.style={
        draw, rounded corners=2pt, minimum height=9mm, minimum width=34mm,
        align=center, fill=purple!12, thick
    },
    decoder/.style={
        draw, rounded corners=2pt, minimum height=10mm, minimum width=46mm,
        align=center, fill=red!12, thick
    },
    output/.style={
        draw, rounded corners=2pt, minimum height=10mm, minimum width=46mm,
        align=center, fill=yellow!20, thick
    },
    grouplabel/.style={font=\footnotesize\itshape, text=gray!70!black},
    arrow/.style={->, thick},
    dasharrow/.style={->, thick, dashed}
]

\node[input] (ctx) at (0,0) {Context set $C = \{(x_c, y_c)\}$};
\node[grouplabel, below=1mm of ctx, align=center]
    {coords + 3$\times$3$\times$128 embedding patch + AGBD};

\node[input, right=28mm of ctx] (tgt) {Target locations $\{x_t\}$};
\node[grouplabel, below=1mm of tgt, align=center]
    {coords + 3$\times$3$\times$128 embedding patch};

\node[encoder, above=12mm of ctx] (cnn_c)
    {CNN Embedding Encoder\\\footnotesize 3$\times$3$\times$128 patch $\to$ 1024-d};
\node[encoder, above=12mm of tgt] (cnn_t)
    {CNN Embedding Encoder\\\footnotesize (shared weights)};

\draw[arrow] (ctx) -- (cnn_c);
\draw[arrow] (tgt) -- (cnn_t);

\node[encoder, above=9mm of cnn_c] (ctx_enc)
    {Context Encoder (MLP)\\\footnotesize $r_c \in \mathbb{R}^{256}$};
\node[smallbox, above=9mm of cnn_t, fill=orange!15, minimum width=34mm] (tgt_q)
    {Target query $\phi(x_t)$};

\draw[arrow] (cnn_c) -- (ctx_enc);
\draw[arrow] (cnn_t) -- (tgt_q);

\node[detbox, above=14mm of ctx_enc] (attn)
    {Multi-Head Cross-Attention (16 heads)\\\footnotesize $Q = \phi(x_t),\;\; K, V = r_c$};

\draw[arrow] (ctx_enc) -- (attn);
\draw[arrow] (tgt_q.north) -- ++(0,4mm) -| (attn.east);

\node[smallbox, above=6mm of attn, fill=green!12, minimum width=34mm] (rdet)
    {Deterministic representation $r_t$};
\draw[arrow] (attn) -- (rdet);

\node[latbox, right=20mm of attn] (pool)
    {Mean Pooling\\\footnotesize aggregate context $r_c$};
\draw[arrow] (ctx_enc.east) to[bend right=20] (pool.south);

\node[latbox, above=6mm of pool] (latent)
    {Latent prior $p(z|C)$\\\footnotesize $\mathcal{N}(\mu_z, \sigma_z)$};
\draw[arrow] (pool) -- (latent);

\node[smallbox, above=6mm of latent, fill=purple!12, minimum width=30mm] (zsample)
    {Sample $z$ \;\footnotesize (reparam.)};
\draw[arrow] (latent) -- (zsample);

\node[decoder, above=16mm of rdet] (dec)
    {Decoder (MLP)\\\footnotesize input: $[\,\phi(x_t),\; r_t,\; z\,]$};

\draw[arrow] (rdet) -- (dec);
\draw[arrow] (zsample.west) to[bend left=15] (dec.east);

\coordinate (bypass) at ($(zsample.east) + (12mm,0)$);
\draw[dasharrow] (tgt_q.east) -- ++(8mm,0) coordinate (bp1)
    -- (bp1 |- bypass) coordinate (bp2)
    -- node[grouplabel, pos=0.5, right]{$\phi(x_t)$} (bp2 |- dec.east)
    -- (dec.east);

\node[output, above=8mm of dec] (out)
    {Predictive distribution $\mathcal{N}(\mu_t, \sigma_t^2)$\\\footnotesize per target location};
\draw[arrow] (dec) -- (out);

\node[grouplabel, above=1mm of out, align=center]
    {mean head $\mu_t$ + log-variance head $\log \sigma_t^2$};

\begin{pgfonlayer}{background}
    \node[draw=orange!70!black, dashed, rounded corners=4pt,
          fit=(cnn_c)(cnn_t)(ctx_enc)(tgt_q),
          inner sep=3mm,
          label={[grouplabel, text=orange!60!black]left:Encoders}] {};

    \node[draw=green!50!black, dashed, rounded corners=4pt,
          fit=(attn)(rdet),
          inner xsep=4mm, inner ysep=3mm,
          label={[grouplabel, text=green!40!black]left:Deterministic Path}] {};

    \node[draw=purple!60!black, dashed, rounded corners=4pt,
          fit=(pool)(latent)(zsample),
          inner sep=5mm,
          label={[grouplabel, text=purple!50!black]right:Stochastic Latent Path}] {};
\end{pgfonlayer}

\end{tikzpicture}}
\caption{Architecture of the Attentive Neural Process (ANP) model.}
\label{fig:anp_architecture}
\end{figure}

\subsection*{Probabilistic Framework}
We specify the biomass estimation problem as learning a distribution over functions. Let a task (or episode) be defined by a set of observed context points $C = \{(x_c, y_c)\}_{c=1}^{N_c}$ and a set of target points $T = \{(x_t, y_t)\}_{t=1}^{N_t}$, where $x \in \mathbb{R}^d$ represents the input features (spatial coordinates and satellite embeddings) and $y \in \mathbb{R}$ represents the biomass density.

We model the predictive distribution $p(y_t | x_t, C)$. The model is trained to maximize the Evidence Lower Bound (ELBO) of the log-likelihood of the target values. The training objective $\mathcal{L}$ is defined as:

\begin{equation}
    \mathcal{L} = \mathbb{E}_{z \sim q(z|C, T)} \left[ \sum_{(x_t, y_t) \in T} \log p(y_t | x_t, r_C, z) \right] - \beta \cdot \text{KL}\left( q(z|C, T) \parallel p(z|C) \right)
\end{equation}

where:
\begin{itemize}
    \item $r_C$ is the deterministic representation of the context set, aggregated via cross-attention.
    \item $z$ is a global latent variable capturing aleatoric uncertainty and spatial correlations not explained by the deterministic path.
    \item $q(z|C, T)$ is the approximate posterior, parameterized as a diagonal Gaussian $\mathcal{N}(\mu_{all}, \sigma_{all})$ encoded from both context and target observations (used during training).
    \item $p(z|C)$ is the conditional prior, parameterized as $\mathcal{N}(\mu_{ctx}, \sigma_{ctx})$ encoded from context observations only. During inference, we sample from this distribution.
    \item $\beta$ is a regularization coefficient. To prevent posterior collapse early in training, we use $\beta$-annealing, linearly increasing $\beta$ over the first 10 epochs.
\end{itemize}

The first term (reconstruction loss) is implemented as the negative log-likelihood of a heteroscedastic Gaussian distribution. The model predicts a mean $\mu_t$ and variance $\sigma^2_t$ for each target point:
\begin{equation}
    \log p(y_t | \cdot) = -\frac{1}{2} \log(2\pi) - \frac{1}{2} \log(\sigma^2_t) - \frac{(y_t - \mu_t)^2}{2\sigma^2_t}
\end{equation}
This allows the model to learn input-dependent (aleatoric) uncertainty, capturing the fact that biomass estimation errors are likely higher in complex, high-biomass forests than in cleared land.

\subsection*{Attention Mechanism}
To capture local spatial dependencies, the deterministic path uses a multihead cross-attention mechanism. For a query point $x_t$, the representation $r_t$ is computed by attending to the context points $(x_c, y_c)$:

\begin{equation}
    r_t = \text{MultiHead}(Q, K, V)
\end{equation}

where the Query ($Q$), Key ($K$), and Value ($V$) are linear projections of the inputs:
\begin{align*}
    Q &= W_q \cdot \phi(x_t) \\
    K &= W_k \cdot \phi(x_c) \\
    V &= W_v \cdot \phi(x_c)
\end{align*}
Here, $\phi(\cdot)$ represents the context encoder network. This mechanism allows the model to weight the contribution of context points based on both spatial proximity and feature similarity, learning a non-linear interpolation kernel.

\subsection*{Training Dynamics and Loss Aggregation}

While standard Neural process implementations usually define the target set as a superset of the context ($C \subseteq T$), we use disjoint split ($C \cap T = \emptyset$) where target points are the complement of context points within a tile. For spatial interpolation tasks, the deployment scenario involves predicting at unobserved locations. Training with disjoint splits better matches this deployment distribution, as the model never predicts at points it has observed. At inference, uncertainty should reflect distance and density of nearby observations. Training on pure interpolation ensures the model learns this spatial structure directly, rather than conflating it with the easier task of reconstructing observed points.

GEDI sampling density varies significantly by tile due to orbital tracks and cloud cover. To prevent densely sampled regions from dominating the gradient, we use tile-averaged loss aggregation. The Negative Log Likelihood (NLL) is first averaged over the target points within each tile, and these scalar tile losses are then averaged across the batch. This weights every spatial unit equally regardless of the number of available GEDI shots, ensuring the model learns to predict robustly in data sparse regions.

\subsection*{Embedding Encoder}

This module processes each $3\times3$ embedding patch, with input shape
$(N, 3, 3, 128)$, into a feature vector. All multilayer perceptrons (MLPs) use ReLU activation functions, and residual connections are applied where noted.

\begin{itemize}
    \item \textbf{Input Projection:} A $1\times1$ convolution maps input channels from $128$ to $256$ to form the residual path.
    
    \item \textbf{Conv Block 1:}  
    $3\times 3$ Conv $(128 \rightarrow 256)$, BatchNorm2d, ReLU.
    
    \item \textbf{Conv Block 2:}  
    $3\times 3$ Conv $(256 \rightarrow 256)$, BatchNorm2d, ReLU, with a residual connection.
    
    \item \textbf{Conv Block 3:}  
    $3\times 3$ Conv $(256 \rightarrow 256)$, BatchNorm2d, ReLU, with a residual connection.
    
    \item \textbf{Pooling:} AdaptiveAvgPool2d to obtain a single feature vector per patch.
    
    \item \textbf{Output MLP:} A 2-layer MLP $(256 \rightarrow 256 \rightarrow 1024)$ with ReLU activation between layers, producing the final embedding feature dimension of $1024$.
\end{itemize}

\subsection*{Context Encoder}

This module encodes each context point (coordinates, embedding features, and AGBD value) into a representation vector.

\begin{itemize}
    \item \textbf{Input Dimension:}  
    $2$ (coords) $+$ $1024$ (embedding features) $+$ $1$ (AGBD)
    $= 1027$.
    
    \item \textbf{Architecture:} A 3-layer MLP with Layer Normalization and residual connections.
    \begin{itemize}
        \item \textbf{Layer 1:} Linear $(1027 \rightarrow 512)$, LayerNorm, ReLU.
        \item \textbf{Layer 2:} Linear $(512 \rightarrow 512)$, LayerNorm, ReLU, with residual connection from Layer 1.
        \item \textbf{Layer 3:} Linear $(512 \rightarrow 512)$, LayerNorm, ReLU, with residual connection from Layer 2.
    \end{itemize}
    
    \item \textbf{Output Projection:} Linear $(512 \rightarrow 256)$, producing the final context representation dimension.
\end{itemize}

\subsection*{Decoder}

This module maps aggregated context information and query features to the final prediction.

\begin{itemize}
    \item \textbf{Input Dimension:}  
    $2$ (coords) $+$ $1024$ (embedding features) $+$ context.
    
    \item \textbf{Architecture:} A 3-layer MLP with Layer Normalization and residual connections.
    \begin{itemize}
        \item \textbf{Layer 1:} Linear $(\text{Input} \rightarrow 512)$, LayerNorm, ReLU.
        \item \textbf{Layer 2:} Linear $(512 \rightarrow 512)$, LayerNorm, ReLU, with residual connection from Layer 1.
        \item \textbf{Layer 3:} Linear $(512 \rightarrow 512)$, LayerNorm, ReLU, with residual connection from Layer 2.
    \end{itemize}
    
    \item \textbf{Output Heads:}
    \begin{itemize}
        \item \textbf{Mean head:} Linear $(512 \rightarrow 1)$.
        \item \textbf{Log-variance head:} Linear $(512 \rightarrow 1)$.
    \end{itemize}
\end{itemize}

\section{Implementation Details}
\label{app:implementation}

\subsection*{Baseline Uncertainty Estimation}

Unlike a Neural Process model, which produces a probabilistic output by design, standard tree-based ensembles require specific adaptation to quantify uncertainty. We implemented the following approaches for the baselines:

\paragraph{Random Forest (Ensemble Variance)}
For Random Forest, we estimated uncertainty using the variance of predictions across individual trees in the ensemble. The predicted standard deviation $\sigma_{pred}$ for a given input $x$ is calculated as:
\[
    \sigma_{pred}(x) = \sqrt{\frac{1}{N_{trees}-1} \sum_{i=1}^{N_{trees}} (\hat{y}_i(x) - \bar{y}(x))^2}
\]
where $\hat{y}_i(x)$ is the prediction of the $i$-th tree. This is the standard method for uncertainty quantification in operational biomass mapping, though it is known to not be an accurate estimate of the true error variance \citep{wager2018estimation}.

\paragraph{XGBoost Quantile Regression}
Standard Gradient Boosting minimizes a squared-error loss, producing point estimates. To generate uncertainty intervals, we trained XGBoost using Quantile Regression. Each quantile model shares the same architecture and hyperparameters as the mean predictor but optimizes the quantile loss. For a chosen quantile level $\alpha$, the models learn
\begin{itemize}
    \item \textbf{Lower Bound:} $q_{1-\alpha}$
    \item \textbf{Upper Bound:} $q_{\alpha}$
\end{itemize}
These quantile predictions form an empirical interval that adapts to heteroscedastic structure in the data. To obtain an approximate predictive standard deviation, we treat the upper-lower quantile span as corresponding to a two-sided 95\% Gaussian interval and compute
\[
\sigma_{\text{pred}} \approx \frac{q_{\alpha} - q_{1-\alpha}}{2 \times 1.96}
\]

\subsection*{Data Processing}
\label{app:preprocessing}

\paragraph{GEDI Quality Filtering}

Using the \texttt{gedidb} library \citep{Besnard2025}, we filter shots based on beam sensitivity, surface detection quality, and agreement with TanDEM-X elevation data. The final training dataset is constructed via an inner join on shot identifiers by using L2A and L2B quality fields (ground-return reliability, canopy cover) to filter L4A footprints. This removes observations in dense canopies where ground detection may be unreliable, even if the L4A product does not explicitly flag them.

\begin{table}[H]
\centering
\caption{GEDI data filtering and merging criteria.}
\label{tab:gedi_filtering}
\begin{tabular}{@{}llp{6.5cm}@{}}
\toprule
\textbf{Product} & \textbf{Filter Variable} & \textbf{Condition / Threshold} \\ \midrule
\textbf{L2A} & \texttt{quality\_flag} & $= 1$ \\
 & \texttt{degrade\_flag} & $= 0$ (Non-degraded) \\
 & \texttt{surface\_flag} & $= 1$ \\
 & \texttt{sensitivity\_a0} & $\in [0.9, 1.0]$ \\
 & \texttt{sensitivity\_a2} & $\in [0.95, 1.0]$ \\
 & \texttt{elevation\_difference\_tdx} & $\in [-150, 150]$ meters (relative to TanDEM-X) \\ \midrule
\textbf{Merge} & Join Key & \texttt{shot\_number} \\
 & Join Type & Inner Join (L2A $\cap$ L2B $\cap$ L4A) \\ \bottomrule
\end{tabular}
\end{table}

\paragraph{Spatial Coordinates}
Longitude and latitude are min-max normalized to the interval $[0, 1]$ using the global bounds of the training set. This preserves the relative spatial geometry while ensuring inputs are in a range suitable for neural networks.

\paragraph{Biomass Density}
Biomass data is positive and approximately log-normally distributed with heteroscedastic noise. We apply a normalized log-transformation to map the data to the $[0, 1]$ interval:
\[
    y' = \frac{\ln(1 + y)}{\ln(1 + S)}
\]
where $S=200$ Mg/ha.

\paragraph{Uncertainty Back-transformation}
The model outputs a predicted standard deviation $\sigma_{norm}$ in the transformed log-space. To convert this back to physical units (Mg/ha), we apply the delta method approximation \citep{Doob1935Limiting} using the derivative of the inverse transformation.
Let $y = \exp(y' \ln(1+S)) - 1$. The standard deviation in the physical space, $\sigma_{raw}$, is related to the normalized output by:
\[
    \sigma_{raw} \approx \sigma_{norm} \cdot \left| \frac{dy}{dy'} \right|
\]
Computing the derivative yields the exact implementation used:
\[
    \sigma_{raw} \approx \sigma_{norm} \cdot \ln(1+S) \cdot (1 + \mu_{raw})
\]
where $\mu_{raw}$ is the predicted mean in Mg/ha. This scales the uncertainty intervals with the magnitude of the biomass, reflecting the log-normal nature of the error distribution.

\paragraph{Numerical Stability Clamping} 
To prevent numerical instability in the probabilistic paths of the model, we constrain the outputs of layers that parameterize distributions.

The decoder module clamps the predicted log-variance to the range $[-7.0, 7.0]$. This prevents the predicted variance from collapsing to zero (log-variance of $-\infty$) or exploding to infinity, both of which can lead to NaN values in the negative log-likelihood loss.

The latent encoder clamps the predicted log standard-deviation to the range $[-10.0, 2.0]$. This stabilizes the calculation of the KL divergence term, particularly by preventing the variance term, $\exp(2 \cdot \log\sigma)$, from becoming excessively large or small.

\paragraph{Gradient Clipping} 
During the training loop, after computing the gradients via backpropagation but before the optimizer step, we apply gradient clipping. We use a maximum norm of 1.0. This mitigates exploding gradients by rescaling gradients whose norm exceeds this threshold, leading to more stable updates and preventing divergence.

\subsection*{Buffered Spatial Cross-Validation}

We implemented a buffered spatial split to reduce spatial dependence between the training, validation, and test sets. First, we randomly selected 15\% of all tiles as test tiles. We then removed any neighboring tiles, defined as lying within $0.1^\circ$ (roughly 11 km) of a test tile. From the remaining pool, we randomly sampled 15\% as validation tiles. The tiles left after formed the final training set. The $0.1^\circ$ buffer distance exceeds typical spatial autocorrelation ranges reported for forest biomass \citep{RejouMechain2014}.

For computational efficiency, we use Euclidean distance in latitude-longitude space:
\[
d(t_1, t_2) = \sqrt{(\text{lon}_2 - \text{lon}_1)^2 + (\text{lat}_2 - \text{lat}_1)^2}
\]

This approximation is valid for our regional study areas of $1^\circ$ squares. For global scale mapping, geodesic distance (Haversine) would be more appropriate to account for Earth's curvature, though the difference is negligible at the spatial scales we consider.

\subsection*{Hyperparameters}
Table \ref{tab:anp_hyperparams} details the hyperparameters used and the training procedure. The model was implemented in PyTorch and trained on a single NVIDIA Tesla T4 GPU.

\begin{table}[H]
    \centering
    \caption{Hyperparameter configuration for the ANP model.}
    \label{tab:anp_hyperparams}
    \begin{tabular}{llc}
        \toprule
        \textbf{Category} & \textbf{Parameter} & \textbf{Value} \\
        \midrule
        \textbf{Architecture} & Hidden Dimension ($d_{model}$) & 512 \\
        & Latent Dimension ($d_z$) & 256 \\
        & Embedding Feature Dim & 1024 \\
        & Attention Heads & 16 \\
        & Context/Target Encoder Layers & 3 \\
        \midrule
        \textbf{Data} & Patch Size & $3 \times 3$ (30m) \\
        & Context Split Ratio & $\mathcal{U}(0.3, 0.7)$ \\
        & Min Shots per Tile & 10 \\
        & Coordinate Noise (Train) & $\mathcal{N}(0, 0.01)$ \\
        \midrule
        \textbf{Training} & Optimizer & AdamW \\
        & Learning Rate & $5 \times 10^{-4}$ \\
        & Weight Decay & $1 \times 10^{-2}$ \\
        & Batch Size & 16 tiles \\
        & Max Epochs & 100 \\
        & LR Scheduler & ReduceLROnPlateau \\
        & Scheduler Factor & 0.5 \\
        & Early Stopping Patience & 15 epochs \\
        \bottomrule
    \end{tabular}
\end{table}

Here we also describe the hyperparameter configurations for the baseline models. For Random Forest and XGBoost, we performed a grid search over the primary complexity parameters as seen in Table~\ref{tab:hyperparams}. Other hyperparameters for these models were held fixed at standard values. For the MLP with MC Dropout, a single, representative configuration was used.

\begin{itemize}
    \item \textbf{Random Forest:} Aside from the swept parameters, all other hyperparameters were left at their default values in the scikit-learn implementation. This includes \texttt{min\_samples\_split=2}, \texttt{min\_samples\_leaf=1}, and \texttt{criterion='squared\_error'}.

    \item \textbf{XGBoost:} The sweep was conducted with the following parameters held constant, based on common practice: \texttt{learning\_rate=0.1}, \texttt{subsample=0.8}, and \texttt{colsample\_bytree=0.8}. Quantile regression models used the \texttt{reg:quantileerror} objective.

    \item \textbf{MLP with MC Dropout:} The MLP baseline was trained using a single configuration. The architecture used a 4-layer MLP with hidden dimensions of (1024, 512, 256, 128) and ReLU activation functions. Training parameters were:
    \begin{itemize}
        \item Optimizer: AdamW
        \item Learning Rate: $5 \times 10^{-4}$
        \item Weight Decay: $1 \times 10^{-5}$
        \item Dropout Rate: 0.2 (applied after each hidden layer)
        \item Batch Size: 256
        \item Epochs: 100 (with early stopping patience of 15 on validation loss)
        \item MC Samples (at inference): 100
    \end{itemize}
\end{itemize}

\subsection*{Region Definitions}
\label{app:region_defs}
For all regions, we aligned the temporal window of the GEDI data with the reference year of the satellite embeddings to minimize errors arising from land cover change (e.g. recent deforestation). For our analysis, we used GEDI observations collected between January 1 and December 31 of 2022, along with foundation model embeddings derived from an annual 2022 composite.

Table \ref{tab:regions} provides the geographic bounding boxes used to filter the GEDI L4A data and query the foundation model embeddings.

\begin{table}[H]
    \centering
    \caption{Geographic specifications and ecological characteristics of the five study regions. Coordinates are provided in WGS84 decimal degrees.}
    \label{tab:regions}
    \hspace*{-1.25cm}
    \begin{tabular}{l l p{5cm} l}
        \toprule
        \textbf{Region Name} & \textbf{Coordinates (Lon / Lat)} & \textbf{Ecological Description} & \textbf{Biome} \\
        \midrule
        \textbf{Guaviare} & $72^\circ$--$73^\circ$W & Tropical lowland forest, & Tropical \\
        (Colombia) & $2^\circ$--$3^\circ$N & deforestation frontier & Moist \\
        \midrule
        \textbf{Tolima} & $74^\circ$--$75^\circ$W & Tropical montane forest, & Tropical \\
        (Colombia) & $3^\circ$--$4^\circ$N & high relief & Montane \\
        \midrule
        \textbf{Maine} & $69^\circ$--$70^\circ$W & Temperate mixed forest, & Temperate \\
        (USA) & $44^\circ$--$45^\circ$N & secondary growth & Broadleaf \\
        \midrule
        \textbf{South Tyrol} & $10.5^\circ$--$11.5^\circ$E & Alpine coniferous forest, & Alpine / \\
        (Italy) & $45.6^\circ$--$46.4^\circ$N & complex terrain & Boreal \\
        \midrule
        \textbf{Hokkaido} & $143.8^\circ$--$144.8^\circ$E & Cool-temperate deciduous & Temperate \\
        (Japan) & $43.2^\circ$--$43.9^\circ$N & and coniferous mix & / Boreal \\
        \bottomrule
    \end{tabular}
\end{table}

We additionally present the summary statistics by region in Table~\ref{tab:region_stats}. The tile count refers to tiles with nonzero data.

\begin{table}[htbp]
\centering
\small
\caption{Summary of GEDI L4A shot statistics by region.}
\label{tab:region_stats}
\begin{tabular}{lccccc}
\toprule
\textbf{Region} &
\textbf{Area (deg$^2$)} &
\textbf{Tiles} &
\textbf{Shots Total} &
\textbf{Mean} &
\textbf{Median} \\
\midrule
Maine (USA) &
1.00 & 100 & 305{,}457 & 3{,}054.6 & 2{,}877 \\
South Tyrol (Italy) &
0.80 & 80 & 337{,}455 & 4{,}218.2 & 3{,}733 \\
Hokkaido (Japan) &
0.70 & 70 & 134{,}564 & 1{,}922.3 & 1{,}628 \\
Tolima (Colombia) &
1.00 & 99 & 82{,}043 & 828.7 & 621 \\
Guaviare (Colombia) &
1.00 & 92 & 144{,}628 & 1{,}572.0 & 1{,}494 \\
\midrule
\toprule
\textbf{Region} &
\textbf{Std. Dev.} &
\textbf{Min} &
\textbf{Max} &
\textbf{Q1} &
\textbf{Q3} \\
\midrule
Maine (USA) &
1{,}376.7 & 277 & 6{,}729 & 1{,}992.3 & 4{,}065.0 \\
South Tyrol (Italy) &
1{,}891.6 & 750 & 8{,}921 & 3{,}057.5 & 5{,}736.3 \\
Hokkaido (Japan) &
1{,}224.4 & 37 & 5{,}169 & 1{,}069.8 & 2{,}784.0 \\
Tolima (Colombia) &
823.3 & 1 & 3{,}470 & 147.0 & 1{,}370.0 \\
Guaviare (Colombia) &
1{,}109.2 & 1 & 4{,}561 & 785.8 & 2{,}163.8 \\
\bottomrule
\end{tabular}
\end{table}

To assess the sensitivity of our results to the AGBD filtering threshold, Table~\ref{tab:filtering_sensitivity} compares dataset characteristics under no filtering, a 500 Mg/ha threshold (used in this study), and a 1000 Mg/ha threshold.

\begin{table}[htbp]
\centering
\small
\caption{Sensitivity of GEDI L4A data to AGBD filtering threshold. For each region, we report the number of retained shots, mean AGBD (Mg/ha), and percentage of shots removed relative to the unfiltered dataset.}
\label{tab:filtering_sensitivity}
\hspace*{-1.25cm}
\begin{tabular}{lcccccc}
\toprule
& \multicolumn{2}{c}{\textbf{Unfiltered}} & \multicolumn{2}{c}{\textbf{$\leq$ 500 Mg/ha}} & \multicolumn{2}{c}{\textbf{$\leq$ 1000 Mg/ha}} \\
\cmidrule(lr){2-3} \cmidrule(lr){4-5} \cmidrule(lr){6-7}
\textbf{Region} & Shots & Mean & Shots & Mean & Shots & Mean \\
\midrule
Maine (USA)          & 305{,}457 & 85.13 & 305{,}046 (0.13\%) & 84.00 & 305{,}326 (0.04\%) & 84.57 \\
South Tyrol (Italy)  & 337{,}455 & 129.41 & 334{,}798 (0.79\%) & 124.65 & 337{,}107 (0.10\%) & 128.10 \\
Hokkaido (Japan)     & 134{,}564 & 93.91 & 134{,}467 (0.07\%) & 93.49 & 134{,}555 (0.01\%) & 93.83 \\
Tolima (Colombia)    & 82{,}043 & 85.05 & 81{,}256 (0.96\%) & 78.97 & 81{,}960 (0.10\%) & 83.73 \\
Guaviare (Colombia)  & 144{,}628 & 55.79 & 144{,}265 (0.25\%) & 54.02 & 144{,}591 (0.03\%) & 55.35 \\
\bottomrule
\end{tabular}
\vspace{0.5em}
\end{table}

\section{Baseline Hyperparameter Sweep}
\label{appendix:hyperparams}

We conducted a grid search over model complexity parameters (Table \ref{tab:hyperparams}) to identify the efficient frontier shown in the main body. We specifically swept \texttt{max\_depth} from 1 (decision stumps) to 20 (deep trees).

\begin{table}[htbp]
\centering
\tiny
\caption{Model hyperparameter sweep for RF and XGB across 5 seed runs sorted by Log $R^2$.}
\begin{tabular}{lcccccc}
\toprule
Model & Max Depth & $n_{\text{estimators}}$ & Test Log $R^2$ & Z-score Std & 1$\sigma$ Coverage (\%) & Train Time (s) \\
\midrule
RF & 20 & 1000 & $0.7147 \pm 0.0262$ & $1.652 \pm 0.428$ & $74.4 \pm 1.8$ & $904.16 \pm 102.58$ \\
RF & 20 & 500 & $0.7144 \pm 0.0262$ & $1.819 \pm 0.537$ & $74.2 \pm 1.7$ & $496.04 \pm 59.42$ \\
RF & 20 & 200 & $0.7136 \pm 0.0266$ & $2.220 \pm 0.891$ & $73.1 \pm 2.0$ & $258.15 \pm 73.03$ \\
RF & 20 & 100 & $0.7124 \pm 0.0262$ & $2.559 \pm 1.022$ & $71.4 \pm 2.4$ & $199.87 \pm 28.59$ \\
RF & 10 & 1000 & $0.7121 \pm 0.0252$ & $3.503 \pm 0.898$ & $75.1 \pm 4.7$ & $539.76 \pm 56.43$ \\
RF & 10 & 500 & $0.7120 \pm 0.0252$ & $3.666 \pm 0.948$ & $74.2 \pm 4.7$ & $281.99 \pm 33.67$ \\
RF & 10 & 200 & $0.7115 \pm 0.0255$ & $3.965 \pm 0.981$ & $72.2 \pm 4.7$ & $139.36 \pm 19.85$ \\
RF & 10 & 50 & $0.7097 \pm 0.0252$ & $4.811 \pm 1.187$ & $65.7 \pm 4.9$ & $104.20 \pm 7.48$ \\
RF & 8 & 1000 & $0.7091 \pm 0.0242$ & $4.399 \pm 0.985$ & $60.2 \pm 5.4$ & $445.08 \pm 51.85$ \\
RF & 8 & 500 & $0.7090 \pm 0.0242$ & $4.547 \pm 1.047$ & $59.7 \pm 5.3$ & $235.37 \pm 32.98$ \\
RF & 8 & 200 & $0.7088 \pm 0.0246$ & $4.838 \pm 1.059$ & $58.3 \pm 5.2$ & $111.91 \pm 13.16$ \\
RF & 8 & 100 & $0.7083 \pm 0.0246$ & $5.251 \pm 1.293$ & $56.5 \pm 4.9$ & $90.17 \pm 11.41$ \\
RF & 8 & 50 & $0.7078 \pm 0.0244$ & $5.498 \pm 1.332$ & $53.6 \pm 4.5$ & $73.58 \pm 6.84$ \\
RF & 6 & 1000 & $0.7018 \pm 0.0235$ & $6.083 \pm 0.913$ & $40.0 \pm 3.7$ & $354.03 \pm 38.50$ \\
RF & 6 & 200 & $0.7017 \pm 0.0236$ & $6.416 \pm 0.893$ & $39.7 \pm 3.7$ & $91.14 \pm 8.45$ \\
RF & 6 & 500 & $0.7017 \pm 0.0235$ & $6.130 \pm 0.890$ & $40.0 \pm 3.6$ & $183.26 \pm 15.60$ \\
RF & 6 & 100 & $0.7015 \pm 0.0236$ & $6.568 \pm 0.982$ & $39.6 \pm 3.5$ & $72.01 \pm 10.53$ \\
RF & 6 & 50 & $0.7010 \pm 0.0237$ & $7.139 \pm 1.079$ & $38.6 \pm 3.4$ & $55.73 \pm 6.41$ \\
RF & 4 & 200 & $0.6858 \pm 0.0234$ & $11.570 \pm 3.510$ & $29.3 \pm 3.8$ & $62.14 \pm 6.38$ \\
RF & 4 & 1000 & $0.6857 \pm 0.0231$ & $11.591 \pm 3.585$ & $29.5 \pm 3.7$ & $249.16 \pm 24.51$ \\
RF & 4 & 500 & $0.6857 \pm 0.0231$ & $11.469 \pm 3.526$ & $29.4 \pm 3.6$ & $138.19 \pm 24.32$ \\
RF & 4 & 100 & $0.6855 \pm 0.0234$ & $12.263 \pm 4.212$ & $29.3 \pm 3.5$ & $51.33 \pm 9.34$ \\
RF & 4 & 50 & $0.6850 \pm 0.0236$ & $14.178 \pm 5.542$ & $28.8 \pm 3.6$ & $37.65 \pm 6.13$ \\
RF & 3 & 1000 & $0.6714 \pm 0.0240$ & $20.274 \pm 7.190$ & $24.3 \pm 3.9$ & $193.60 \pm 21.02$ \\
RF & 3 & 500 & $0.6713 \pm 0.0239$ & $20.379 \pm 7.162$ & $24.3 \pm 3.7$ & $101.26 \pm 12.42$ \\
RF & 3 & 200 & $0.6713 \pm 0.0243$ & $21.256 \pm 7.830$ & $24.1 \pm 3.9$ & $46.97 \pm 4.58$ \\
RF & 3 & 100 & $0.6711 \pm 0.0242$ & $22.288 \pm 7.586$ & $23.9 \pm 3.8$ & $37.91 \pm 5.22$ \\
RF & 3 & 50 & $0.6706 \pm 0.0243$ & $23.505 \pm 8.467$ & $23.1 \pm 3.5$ & $29.14 \pm 3.22$ \\
RF & 2 & 200 & $0.6395 \pm 0.0254$ & $35.215 \pm 6.723$ & $16.4 \pm 1.8$ & $33.15 \pm 3.54$ \\
RF & 2 & 1000 & $0.6394 \pm 0.0251$ & $33.826 \pm 7.237$ & $16.5 \pm 1.9$ & $134.54 \pm 15.22$ \\
RF & 2 & 500 & $0.6393 \pm 0.0250$ & $33.879 \pm 7.137$ & $16.3 \pm 1.9$ & $71.36 \pm 8.30$ \\
RF & 2 & 100 & $0.6392 \pm 0.0254$ & $35.758 \pm 5.427$ & $16.3 \pm 1.4$ & $25.65 \pm 2.82$ \\
RF & 2 & 50 & $0.6384 \pm 0.0254$ & $37.169 \pm 5.920$ & $15.8 \pm 1.0$ & $20.70 \pm 3.04$ \\
RF & 1 & 1000 & $0.5632 \pm 0.0305$ & $61.624 \pm 23.731$ & $6.6 \pm 5.2$ & $71.94 \pm 7.80$ \\
RF & 1 & 500 & $0.5632 \pm 0.0307$ & $64.325 \pm 24.967$ & $6.5 \pm 5.0$ & $37.56 \pm 4.49$ \\
RF & 1 & 200 & $0.5619 \pm 0.0326$ & $64.077 \pm 21.728$ & $6.1 \pm 4.0$ & $17.77 \pm 1.73$ \\
RF & 1 & 100 & $0.5611 \pm 0.0334$ & $65.387 \pm 23.265$ & $6.0 \pm 4.0$ & $13.12 \pm 1.26$ \\
RF & 1 & 50 & $0.5583 \pm 0.0337$ & $69.941 \pm 24.866$ & $5.3 \pm 3.8$ & $10.33 \pm 1.01$ \\
\midrule
XGB & 6 & 100 & $0.7177 \pm 0.0244$ & $1.444 \pm 0.076$ & $65.5 \pm 1.6$ & $123.79 \pm 2.13$ \\
XGB & 6 & 50 & $0.7175 \pm 0.0244$ & $1.373 \pm 0.079$ & $67.4 \pm 1.6$ & $65.19 \pm 1.39$ \\
XGB & 4 & 200 & $0.7174 \pm 0.0242$ & $1.447 \pm 0.069$ & $65.7 \pm 1.8$ & $180.59 \pm 3.51$ \\
XGB & 4 & 100 & $0.7166 \pm 0.0244$ & $1.377 \pm 0.065$ & $68.1 \pm 1.7$ & $93.42 \pm 1.67$ \\
XGB & 3 & 500 & $0.7166 \pm 0.0247$ & $1.475 \pm 0.066$ & $64.5 \pm 2.2$ & $367.55 \pm 1.62$ \\
XGB & 6 & 200 & $0.7165 \pm 0.0251$ & $1.531 \pm 0.070$ & $62.7 \pm 1.5$ & $241.67 \pm 3.45$ \\
XGB & 4 & 500 & $0.7161 \pm 0.0251$ & $1.562 \pm 0.062$ & $62.1 \pm 2.0$ & $443.18 \pm 6.42$ \\
XGB & 8 & 50 & $0.7161 \pm 0.0253$ & $1.390 \pm 0.066$ & $66.0 \pm 1.6$ & $83.57 \pm 1.02$ \\
XGB & 3 & 200 & $0.7161 \pm 0.0245$ & $1.400 \pm 0.074$ & $67.1 \pm 2.3$ & $150.86 \pm 2.28$ \\
XGB & 2 & 1000 & $0.7159 \pm 0.0248$ & $1.533 \pm 0.069$ & $64.0 \pm 1.6$ & $578.00 \pm 7.82$ \\
XGB & 2 & 500 & $0.7153 \pm 0.0246$ & $1.427 \pm 0.052$ & $66.0 \pm 2.0$ & $292.22 \pm 2.41$ \\
XGB & 3 & 1000 & $0.7151 \pm 0.0246$ & $1.572 \pm 0.062$ & $61.6 \pm 1.3$ & $734.55 \pm 6.81$ \\
XGB & 8 & 100 & $0.7149 \pm 0.0257$ & $1.492 \pm 0.068$ & $63.1 \pm 1.3$ & $159.71 \pm 2.15$ \\
XGB & 4 & 50 & $0.7140 \pm 0.0244$ & $1.303 \pm 0.054$ & $70.0 \pm 1.8$ & $49.29 \pm 1.18$ \\
XGB & 1 & 1000 & $0.7128 \pm 0.0244$ & $1.125 \pm 0.033$ & $76.5 \pm 2.3$ & $428.93 \pm 5.97$ \\
XGB & 6 & 500 & $0.7128 \pm 0.0256$ & $1.805 \pm 0.154$ & $56.8 \pm 1.3$ & $596.01 \pm 4.66$ \\
XGB & 2 & 200 & $0.7120 \pm 0.0244$ & $1.339 \pm 0.049$ & $68.6 \pm 2.3$ & $118.45 \pm 2.35$ \\
XGB & 10 & 50 & $0.7120 \pm 0.0266$ & $1.438 \pm 0.077$ & $63.5 \pm 1.6$ & $109.75 \pm 1.32$ \\
XGB & 3 & 50 & $0.7096 \pm 0.0245$ & $1.269 \pm 0.040$ & $70.4 \pm 2.1$ & $40.96 \pm 0.82$ \\
XGB & 10 & 100 & $0.7109 \pm 0.0265$ & $1.563 \pm 0.089$ & $60.1 \pm 1.3$ & $205.03 \pm 2.45$ \\
XGB & 8 & 500 & $0.7109 \pm 0.0259$ & $2.011 \pm 0.207$ & $53.1 \pm 1.7$ & $765.95 \pm 9.95$ \\
XGB & 10 & 200 & $0.7098 \pm 0.0267$ & $1.749 \pm 0.135$ & $55.9 \pm 1.3$ & $391.56 \pm 4.52$ \\
XGB & 6 & 1000 & $0.7099 \pm 0.0260$ & $2.865 \pm 1.285$ & $51.9 \pm 1.4$ & $1184.81 \pm 14.78$ \\
XGB & 10 & 500 & $0.7092 \pm 0.0267$ & $2.533 \pm 0.491$ & $50.3 \pm 1.6$ & $946.80 \pm 4.91$ \\
XGB & 10 & 1000 & $0.7091 \pm 0.0267$ & $153.148 \pm 263.490$ & $45.9 \pm 1.7$ & $1872.18 \pm 37.74$ \\
XGB & 1 & 500 & $0.7088 \pm 0.0242$ & $1.110 \pm 0.030$ & $77.1 \pm 2.3$ & $216.72 \pm 3.59$ \\
XGB & 2 & 100 & $0.7081 \pm 0.0243$ & $1.273 \pm 0.034$ & $70.5 \pm 2.6$ & $60.80 \pm 1.22$ \\
XGB & 8 & 1000 & $0.7104 \pm 0.0259$ & $40.237 \pm 71.707$ & $48.7 \pm 1.7$ & $1510.02 \pm 16.14$ \\
XGB & 2 & 50 & $0.7022 \pm 0.0243$ & $1.207 \pm 0.029$ & $71.9 \pm 2.5$ & $31.83 \pm 0.69$ \\
XGB & 1 & 200 & $0.7022 \pm 0.0244$ & $1.087 \pm 0.020$ & $77.8 \pm 2.0$ & $88.07 \pm 1.55$ \\
XGB & 20 & 100 & $0.7021 \pm 0.0264$ & $1.879 \pm 0.168$ & $52.1 \pm 1.9$ & $659.32 \pm 30.14$ \\
XGB & 20 & 200 & $0.7021 \pm 0.0263$ & $8.233 \pm 6.825$ & $46.8 \pm 2.2$ & $1030.88 \pm 38.61$ \\
XGB & 20 & 500 & $0.7020 \pm 0.0263$ & $35.153 \pm 14.312$ & $40.2 \pm 2.5$ & $2066.57 \pm 62.07$ \\
XGB & 20 & 1000 & $0.7020 \pm 0.0264$ & $222.272 \pm 208.624$ & $33.9 \pm 2.6$ & $3721.61 \pm 110.13$ \\
XGB & 20 & 50 & $0.7017 \pm 0.0264$ & $1.605 \pm 0.112$ & $56.1 \pm 1.9$ & $397.20 \pm 21.34$ \\
XGB & 1 & 100 & $0.6972 \pm 0.0245$ & $1.067 \pm 0.021$ & $78.1 \pm 2.0$ & $45.40 \pm 0.65$ \\
XGB & 1 & 50 & $0.6888 \pm 0.0238$ & $1.038 \pm 0.025$ & $78.2 \pm 1.8$ & $23.79 \pm 0.42$ \\
\bottomrule
\end{tabular}
\label{tab:hyperparams}
\end{table}

\section{NP Architecture Ablation}
\label{appendix:ablation}

Are the additional complexities of the ANP model justified over a simpler neural process? We conducted an ablation over model architecture of the model (Table \ref{tab:ablation}). We specifically ablated the deterministic attention component, as well as the stochastic latent component, over 5 seeds in the Guaviare region of interest defined in Table~\ref{tab:regions}.

The ablation result suggests a secondary regularization role for the latent path, even if the latent-alone model performance is not competitive. The full ANP exhibits lower variance in its performance metrics across runs compared to the more volatile deterministic-only model. This suggests that the stochastic latent path, by forcing the model to learn a compressed, global summary of the landscape, acts as a variational bottleneck. This prevents the attention-based deterministic path from overfitting to spurious local patterns in the training data, leading to more stable and generalizable solutions.

\begin{table}[H]
\centering
\caption{Ablation on the components of the ANP architecture.}
\begin{tabular}{lccc}
\toprule
\textbf{Metric} & \textbf{Full} & \textbf{Deterministic} & \textbf{Latent} \\
\midrule
\multicolumn{4}{l}{\textit{Accuracy Metrics}} \\
\quad Log-space $R^2$ & 0.722 $\pm$ 0.017 & 0.715 $\pm$ 0.032 & 0.699 $\pm$ 0.045 \\
\quad Log-space MAE & 0.146 $\pm$ 0.011 & 0.151 $\pm$ 0.009 & 0.147 $\pm$ 0.010 \\
\quad Linear RMSE (Mg/ha) & 55.85 $\pm$ 5.14 & 54.59 $\pm$ 1.75 & 53.94 $\pm$ 2.30 \\
\midrule
\multicolumn{4}{l}{\textit{Uncertainty Calibration}} \\
\quad Z-score Std (Ideal: 1.0) & 1.03 $\pm$ 0.06 & 1.05 $\pm$ 0.14 & 1.05 $\pm$ 0.07 \\
\quad Z-score Mean (Ideal: 0.0) & 0.06 $\pm$ 0.02 & -0.01 $\pm$ 0.11 & -0.04 $\pm$ 0.14 \\
\quad 1$\sigma$ Coverage (\%) & 76.8 $\pm$ 4.2 & 76.4 $\pm$ 5.2 & 78.1 $\pm$ 2.5 \\
\bottomrule
\end{tabular}
\label{tab:ablation}
\end{table}

\section{Regression Kriging}
\label{appendix:kriging}

What about classical geostatistics? To provide a comparison, we implemented a geostatistical baseline with Regression Kriging (RK), or Gaussian Process Regression. This method combines a random forest for the global trend with a geostatistical model for the spatially autocorrelated residuals.

Our RK implementation follows a standard procedure:
\begin{enumerate}
    \item \textbf{Trend Modeling:} We first fit a Random Forest, identical to the RF baseline in the main body.
    \item \textbf{Residual Kriging:} We then computed the residuals from the RF model's predictions on the training set. A variogram was fitted to these residuals to model their spatial autocorrelation structure. Ordinary Kriging was applied to these residuals to produce a spatially continuous surface of expected residual values.
\end{enumerate}
The final prediction at a target location \(x^*\) is the sum of the RF trend prediction and the kriged residual prediction:
\[
\text{AGB}_{\text{RK}}(x^*) = \text{AGB}_{\text{RF}}(x^*) + \text{Residual}_{\text{Krige}}(x^*)
\]
The predictive uncertainty is derived from the kriging variance, which adapts to the local density of training data points. All experiments were conducted on the Guaviare study region using the same 10 seeds and buffered spatial splits as the primary analysis.

We evaluated the model on the same metrics as the main baselines. The aggregated results are presented in Table~\ref{tab:rk_comparison}, alongside the results for the ANP and RF models from the main body for comparison.

\begin{table}[H]
\centering
\caption{Performance comparison of Regression Kriging against Attentive Neural Process and Random Forest on the Guaviare test set.}
\small
\label{tab:rk_comparison}
\begin{tabular}{lccc}
\toprule
Model & Regression Kriging & ANP & Random Forest \\
\midrule
\multicolumn{4}{l}{\textit{Accuracy Metrics}} \\
Test Log $R^2$ & $0.723 \pm 0.044$ & \textbf{$0.747 \pm 0.043$} & $0.724 \pm 0.045$ \\
Test Log RMSE & $0.208 \pm 0.014$ & \textbf{$0.199 \pm 0.016$} & $0.207 \pm 0.015$ \\
Test Log MAE & $0.149 \pm 0.010$ & \textbf{$0.141 \pm 0.012$} & $0.148 \pm 0.010$ \\
Linear RMSE (Mg/ha) & $51.65 \pm 3.58$ & \textbf{$50.56 \pm 5.36$} & $51.68 \pm 3.52$ \\
Linear MAE (Mg/ha) & $28.00 \pm 2.82$ & \textbf{$27.33 \pm 3.44$} & $28.01 \pm 2.71$ \\
\midrule
\multicolumn{4}{l}{\textit{Uncertainty Calibration}} \\
Z-Score Mean (0.0) & $-0.012 \pm 0.054$ & $0.023 \pm 0.124$ & $-0.109 \pm 0.407$ \\
Z-Score Std (1.0) & \textbf{$0.984 \pm 0.073$} & $0.997 \pm 0.099$ & $6.960 \pm 2.065$ \\
1$\sigma$ Coverage (68.3\%) & $79.4 \pm 2.5$ & $79.1 \pm 2.8$ & $19.1 \pm 3.3$ \\
2$\sigma$ Coverage (95.4\%) & $94.1 \pm 1.3$ & $94.1 \pm 1.4$ & $39.8 \pm 5.0$ \\
3$\sigma$ Coverage (99.7\%) & $98.1 \pm 0.8$ & $98.0 \pm 0.9$ & $59.8 \pm 5.2$ \\
\bottomrule
\end{tabular}
\end{table}

Regression Kriging achieves good uncertainty calibration, with a Z-score standard deviation of 0.984. This confirms that explicitly modeling the spatial structure of model residuals is an effective strategy for producing well-calibrated prediction intervals. The calibration quality is superior to the uncalibrated ensemble variance from the standalone random forest model.

However, regression kriging offers no significant improvement in predictive accuracy over a standard Random Forest. It is virtually identical to that of the plain random forest and underperforms the ANP. This suggests that while kriging the residuals correctly accounts for spatial proximity, it does not capture the more complex, feature dependent spatial representations that the ANP's attention mechanism can learn.

The primary limitation of regression kriging is its computational inefficiency ($O(N^3)$) and memory complexity ($O(N^2)$). With a mean training time of over 400 seconds for a single 1x1 degree area and high variance (std 302s) depending on data density, the method is approximately 3 to 6 times slower than the ANP. The cost of variogram fitting and kriging scales poorly with the number of data points, rendering this approach intractable for the large area, high resolution mapping tasks that are the focus of this work. Modern sparse GP approximations with inducing points reduce it to $O(M)+O(M^2)$ per prediction, but still incur non-trivial per-prediction cost that scales with the number of inducing points, and the quality of the approximation depends on inducing point optimization, which is itself a nontrivial problem over heterogeneous landscapes.

ANPs amortize this cost as the training learns a function that directly maps (context set, target location) to predictive distributions. Inference is a single forward pass with fixed computational cost regardless of context size. For continental-scale mapping where predictions are needed at 10m resolution, this distinction between fast training, slow inference (sparse GPs) and moderate training, essentially free inference (ANPs) becomes operationally decisive. For this reason, we did not include it as a standard baseline in the subsequent experiments of the main body.

\section{Quantile Random Forest}
\label{appendix:qrf}

We also tested a Quantile Random Forest (QRF) baseline. Unlike a standard Random Forest that estimates the mean, a QRF is designed to approximate the full conditional distribution of the target variable, enabling the calculation of prediction intervals from quantiles.

Our implementation used the \texttt{quantile-forest} Python package. To ensure a fair comparison, the QRF was configured with the same hyperparameters as the representative Random Forest model in the main body (e.g. number of trees, max depth). The experiments were conducted on the Guaviare study region using the identical 10 seeds and buffered spatial splits.

The aggregated results are presented in Table~\ref{tab:qrf_comparison}, alongside the results for the ANP and the standard RF models from the main body for context.

\begin{table}[H]
\centering
\caption{Performance comparison of Quantile Random Forest (QRF) against the Attentive Neural Process and Random Forest on the Guaviare test set.}
\small
\label{tab:qrf_comparison}
\begin{tabular}{lccc}
\toprule
Model & Quantile Random Forest & ANP & Random Forest \\
\midrule
\multicolumn{4}{l}{\textit{Accuracy Metrics}} \\
Test Log $R^2$ & $0.688 \pm 0.057$ & \textbf{$0.747 \pm 0.043$} & $0.724 \pm 0.045$ \\
Test Log RMSE & $0.220 \pm 0.018$ & \textbf{$0.199 \pm 0.016$} & $0.207 \pm 0.015$ \\
Test Log MAE & $0.130 \pm 0.012$ & $0.141 \pm 0.012$ & $0.148 \pm 0.010$ \\
Linear RMSE (Mg/ha) & $49.03 \pm 3.57$ & \textbf{$50.56 \pm 5.36$} & $51.68 \pm 3.52$ \\
Linear MAE (Mg/ha) & $27.06 \pm 2.79$ & \textbf{$27.33 \pm 3.44$} & $28.01 \pm 2.71$ \\
\midrule
\multicolumn{4}{l}{\textit{Uncertainty Calibration}} \\
Z-Score Mean (0.0) & $0.130 \pm 0.081$ & $0.023 \pm 0.124$ & $-0.109 \pm 0.407$ \\
Z-Score Std (1.0) & $1.142 \pm 0.122$ & \textbf{$0.997 \pm 0.099$} & $6.960 \pm 2.065$ \\
1$\sigma$ Coverage (68.3\%) & $77.2 \pm 2.7$ & $79.1 \pm 2.8$ & $19.1 \pm 3.3$ \\
2$\sigma$ Coverage (95.4\%) & $91.4 \pm 1.6$ & \textbf{$94.1 \pm 1.4$} & $39.8 \pm 5.0$ \\
3$\sigma$ Coverage (99.7\%) & $96.6 \pm 1.1$ & \textbf{$98.0 \pm 0.9$} & $59.8 \pm 5.2$ \\
\bottomrule
\end{tabular}
\end{table}

The QRF successfully rectifies the miscalibration of the standard RF's ensemble variance. With a Z-score standard deviation of 1.14, it is superior to the standard RF (6.96) and approaches the near-ideal calibration of the ANP (0.997). This confirms that a quantile based approach provides a more reliable foundation for uncertainty estimation than naive ensemble variance.

However, this improvement in calibration comes at a significant cost to predictive accuracy. The QRF's mean test Log $R^2$ of 0.688 is substantially lower than both the standard RF (0.724) and the ANP (0.747). This accuracy drop off arises because the QRF must expend its model capacity on approximating the entire conditional distribution, rather than focusing solely on minimizing the error of the mean estimate.

Compared to the XGBoost quantile regression baseline (Log $R^2$ 0.737, Z-std 1.408), the QRF presents an unfavorable trade off. It offers a modest improvement in calibration for a large drop in accuracy. Given this performance profile, the QRF does not represent a competitive alternative to either the ANP or even the more pragmatic XGBoost QR baseline. For this reason, we did not include QRF in the main body of the paper or further experiments.

\section{Deep Ensemble}
\label{appendix:deep_ensemble}

To isolate the contribution of the ANP's context-conditioning architecture from its heteroscedastic likelihood, we also test a deep ensemble baseline \citep{Lakshminarayanan2017scalable}. Deep ensembles are widely regarded as a strong baseline for uncertainty quantification in neural networks, combining epistemic uncertainty (disagreement across independently trained members) with, optionally, aleatoric uncertainty (per-member variance prediction).

We train ensembles of 5 independently initialized MLPs sharing the same architecture as the MC Dropout baseline. We evaluate two variants, (1) a standard ensemble trained with MSE loss, where uncertainty is derived from the variance of member predictions, and (2) a deep ensemble trained with Gaussian negative log-likelihood loss, where each member predicts both mean $\mu$ and variance $\sigma^2$, and the predictive distribution is a Gaussian mixture combined via the law of total variance:
\begin{equation}
    \sigma^2_{\text{mix}} = \frac{1}{M}\sum_{m=1}^{M}(\sigma_m^2 + \mu_m^2) - \mu_{\text{mix}}^2
\end{equation}
where $\mu_{\text{mix}} = \frac{1}{M}\sum_{m=1}^{M}\mu_m$. All experiments were conducted on the Guaviare study region using the same 10 seeds and buffered spatial splits as the primary analysis.

\begin{table}[H]
\centering
\caption{Performance comparison of deep ensemble variants against the Attentive Neural Process and MC Dropout baseline on the Guaviare test set.}
\small
\label{tab:ensemble_comparison}
\hspace*{-1.5cm}
\begin{tabular}{lcccc}
\toprule
Model & Deep Ens. (NLL) & Deep Ens. (MSE) & ANP & MLP w/ Dropout \\
\midrule
\multicolumn{5}{l}{\textit{Accuracy Metrics}} \\
Test Log $R^2$ & $0.734 \pm 0.049$ & $0.712 \pm 0.055$ & $\textbf{0.747} \pm 0.043$ & $0.709 \pm 0.053$ \\
Test Log RMSE & $0.204 \pm 0.017$ & $0.212 \pm 0.018$ & $\textbf{0.199} \pm 0.016$ & $0.213 \pm 0.017$ \\
Linear RMSE (Mg/ha) & $\textbf{48.96} \pm 2.58$ & $48.98 \pm 3.14$ & $50.56 \pm 5.36$ & $48.71 \pm 2.84$ \\
\midrule
\multicolumn{5}{l}{\textit{Uncertainty Calibration}} \\
Z-Score Mean (0.0) & $0.034 \pm 0.087$ & $-0.041 \pm 0.112$ & $\textbf{0.023} \pm 0.124$ & $0.041 \pm 0.249$ \\
Z-Score Std (1.0) & $1.757 \pm 0.389$ & $10.257 \pm 2.479$ & $\textbf{0.997} \pm 0.099$ & $4.543 \pm 0.472$ \\
1$\sigma$ Coverage (68.3\%) & $71.4 \pm 2.4$ & $41.8 \pm 2.3$ & $\textbf{79.1} \pm 2.8$ & $24.1 \pm 4.2$ \\
2$\sigma$ Coverage (95.4\%) & $90.5 \pm 1.6$ & $69.8 \pm 2.4$ & $\textbf{94.1} \pm 1.4$ & $52.8 \pm 7.5$ \\
3$\sigma$ Coverage (99.7\%) & $96.5 \pm 1.1$ & $82.3 \pm 2.2$ & $\textbf{98.0} \pm 0.9$ & $73.1 \pm 3.7$ \\
\midrule
\multicolumn{5}{l}{\textit{Computational Cost}} \\
Train Time (s) & $1435 \pm 480$ & $928 \pm 245$ & $\textbf{146} \pm 12$ & $185 \pm 57$ \\
\bottomrule
\end{tabular}
\end{table}

The comparison between the two ensemble variants isolates the effect of the loss function. The MSE ensemble (Z-std 10.26, 1$\sigma$ coverage 41.8\%) performs comparably to the Random Forest baseline from the main body (Z-std 6.96, coverage 19.1\%), confirming that ensemble disagreement alone even from independently trained neural networks does not produce calibrated prediction intervals. The heteroscedastic ensemble substantially improves calibration (Z-std 1.76, coverage 71.4\%), demonstrating that training with Gaussian NLL loss is necessary for reasonable uncertainty estimates.

However, the heteroscedastic ensemble remains overconfident relative to the ANP (Z-std 0.997, coverage 79.1\%). This residual miscalibration reflects the architectural difference, since each ensemble member predicts independently from the target location's features, without conditioning on the set of nearby GEDI observations. The ensemble can learn that certain feature combinations are harder to predict (higher aleatoric $\sigma$) and that some regions of feature space are underrepresented (higher epistemic spread), but it cannot observe that nearby context points scatter and widen its intervals accordingly.

The computational cost further differentiates the approaches. The heteroscedastic ensemble requires training 5 independent networks, resulting in approximately 10$\times$ the training time of the ANP (1435s vs 146s) for inferior calibration. This cost scales linearly with ensemble size, whereas the ANP's context-conditioning is amortized during training and adds negligible inference cost.

These results confirm that heteroscedastic likelihood training is necessary but not sufficient for well-calibrated biomass prediction intervals. The ANP achieves superior calibration at a fraction of the computational cost. Given that the heteroscedastic ensemble achieves inferior calibration to the ANP at approximately 10 times the computational cost and does not clearly dominate ANP on any axis, we do not extend this comparison to the additional study regions in Section~\ref{sec:multiregion}.

\end{document}